\documentclass[11pt]{article}

\usepackage[final]{acl}

\usepackage{microtype}

\usepackage[english,bidi=default]{babel} 

\usepackage{times}
\usepackage{latexsym}

\usepackage[T1]{fontenc}

\usepackage[utf8]{inputenc}

\usepackage{microtype}

\usepackage{inconsolata}

\usepackage{graphicx}
\usepackage{booktabs}
\usepackage{multirow}
\usepackage{subfigure}
\usepackage{amsmath}
\usepackage{amssymb}
\usepackage{hyperref}
\usepackage{float}
\usepackage{multirow}
\usepackage{booktabs}
\usepackage{amsmath}
\usepackage{wrapfig}
\usepackage{pifont}

\usepackage[ruled,vlined]{algorithm2e}

\usepackage{arydshln}

\newcommand{\METHODNAME}{\textsc{GenesisFunc}}

\title{\METHODNAME{}: Multi-Agent Data Generation for Accurate and Generalizable Function-Calling}

\author{
Hao-Xiang Xu, Chong Deng$^1$, Jiaqing Liu$^1$, Wen Wang$^1$, Qian Chen$^1$,\\
\textbf{Lujia Bao}$^1$\textbf{,} \textbf{Xiangang Li}$^1$\textbf{,} \textbf{Zhen-Hua Ling}\thanks{Corresponding author.} \\
$^1$Tongyi Fun Team, Alibaba Group \\
{\tt nh2001620@mail.ustc.edu.cn}
}

\begin{document}

\maketitle
\begin{abstract}
Large Language Models (LLMs) extend their capabilities through function-calling (FC), which relies on training data with high quality, diversity, and broad coverage of scenarios. However, obtaining and annotating real function-calling data is challenging, while synthetic data from existing pipelines often suffers from unreliable APIs, limited tool scalability, insufficient diversity, and weak quality control.
To address these, we present \METHODNAME{}, an automated pipeline for generating FC training data. Starting from reliable tools in widely used public benchmarks, our \METHODNAME{} employs a multi-agent framework to support a dialogue generation system that produces conversations spanning diverse scenarios, while maintaining both diversity and quality throughout the process. The accuracy of the data is further reinforced through a multi-stage evaluation system.
We fine-tune an 8B LLM on the synthetic dataset and show through extensive experiments that it outperforms similarly sized open-source models in in-domain FC performance and out-of-domain generalization, while reaching FC capabilities comparable to some of the latest API-based models. In addition, our method demonstrates strong potential to scale effectively across downstream tools, underscoring its real-world applicability.
The complete pipeline and the constructed dataset is available at \href{https://github.com/famoustourist/GenesisFunc}{https://github.com/famoustourist/GenesisFunc}.
\end{abstract}

\section{Introduction}
Tool learning represents a crucial step in advancing the frontier of Large Language Models (LLMs), as it transforms them from passive language processors into proactive agents capable of interacting with dynamic environments~\cite{DBLP:conf/acl/HuangZLZGLHZWSJ24, DBLP:conf/iclr/QinLYZYLLCTQZHT24}. By bridging internal reasoning with external execution, tool-equipped LLMs are no longer constrained by static training data but can generalize to open-ended tasks and adapt to evolving user needs~\cite{DBLP:conf/nips/PatilZ0G24, DBLP:journals/fcsc/QuDWCWYXW25}. This paradigm thus underscores not only the practical value of enhancing task coverage across diverse applications such as workflow automation and travel planning~\cite{DBLP:journals/corr/abs-2401-12224, DBLP:journals/corr/abs-2404-11891}, but also the theoretical importance of extending the fundamental boundaries of LLM capabilities~\cite{DBLP:conf/nips/LiuHZZLKTYLFNYS24, DBLP:conf/iclr/Liu0ZHYL0GLY0WN25}.

Despite the rapid progress in tool learning, the field still faces fundamental challenges that constrain its broader applicability. The effectiveness of function-calling (FC) relies heavily on high-quality training data, yet collecting and annotating real-world data is costly and labor-intensive~\cite{DBLP:conf/iclr/QinLYZYLLCTQZHT24}. Moreover, real-world function-calls are inherently complex and diverse, often characterized by ambiguous user intents, rapidly changing dynamic environments, multi-task requirements, and extended multi-turn interactions~\cite{DBLP:journals/corr/abs-2306-05301, DBLP:conf/nips/PatilZ0G24, DBLP:conf/nips/LiuHZZLKTYLFNYS24, DBLP:conf/emnlp/Abdelaziz0AKSPR24}. These intertwined challenges underscore the urgent need for more robust data generation pipelines that can produce \textit{accurate}, \textit{diverse}, and \textit{broadly representative training data} to support realistic function-calling scenarios.

Increasing research efforts have sought to design automated pipelines for synthesizing training data, which are then used to build tool-augmented LLMs~\cite{DBLP:journals/fcsc/QuDWCWYXW25, DBLP:conf/iclr/Liu0ZHYL0GLY0WN25}. While progress has been made, fundamental limitations remain. Existing methods often rely on publicly available or manually constructed APIs, which are frequently unreliable and difficult to scale across broader tool sets, thereby constraining function-calling capabilities. In addition, the generated data often suffers from insufficient diversity and inadequate quality control, further weakening performance. From the perspective of scenario coverage, most approaches still emphasize single-turn or isolated function calls, leaving them unable to capture more realistic settings such as multi-turn dialogues or multi-task interactions. Together, these limitations suggest the insufficiency of current approaches and highlight the need for more effective and scalable solutions.

In this paper, we introduce \textbf{\METHODNAME{}}, a three-stage automated pipeline designed to generate function-calling training data that is \textbf{high-quality}, \textbf{diverse}, and \textbf{broadly representative of real-world scenarios}.
To ensure the reliability of APIs and enable seamless extension to a wider range of tools, \METHODNAME{} begins by \textbf{selecting a curated seed set} from the widely adopted BFCL benchmark~\cite{BFCL}. These APIs are verified to cover multiple domains, making them not only dependable but also practical for extension to downstream tools, since such functions are readily accessible in real applications~\cite{DBLP:journals/corr/abs-2402-10891}.
Next, to improve dialogue quality, diversity, and coverage, \METHODNAME{} incorporates a \textbf{multi-agent-assisted dialogue generation system}. Through the coordinated interaction of agents, the framework leverages history-aware role differentiation and parameter-slot selection to expand diversity, while an agent-based scoring mechanism safeguards the quality of synthesized dialogues.
Finally, to further guarantee correctness, \METHODNAME{} employs a \textbf{multi-stage evaluation module} that combines rule-based and model-based checks, followed by targeted human review, thereby ensuring both conformity and executability of the training data.
Through this end-to-end process, \METHODNAME{} produces robust, diverse, and scenario-rich training data, which significantly enhances the function-calling capabilities of LLMs.

To validate the effectiveness of \METHODNAME{}, we conduct supervised fine-tuning on Qwen3-8B~\cite{DBLP:journals/corr/abs-2505-09388} using the training data generated by our pipeline and evaluate the zero-shot performance of the fine-tuned model across several public benchmarks. Since the tools are sourced from BFCL~\cite{BFCL}, we report results not only on the in-domain BFCL benchmark but also on the out-of-domain API-Bank~\cite{DBLP:conf/emnlp/LiZ000YLHL23} and ACEBench~\cite{DBLP:journals/corr/abs-2501-12851}, thereby assessing both in-domain performance and generalization to unseen domains. Under a consistent evaluation protocol, our model consistently surpasses open-source baselines of comparable scales and remains highly competitive with API-based models~\cite{DBLP:journals/corr/abs-2303-08774}. Furthermore, we demonstrate that our approach can scale up to more downstream tools and verify that the data generated by our pipeline can also be used through reinforcement learning to enhance function-calling performance in multi-turn dialogue scenarios.

In summary, this paper makes three primary contributions:
(1) We introduce \METHODNAME{}, an automated pipeline for generating high-quality function-calling training data for LLMs. The pipeline begins with selecting reliable APIs and integrates a \textbf{Multi-agent Dialogue Generation Module} and a \textbf{Multi-stage Evaluation Module} to ensure robustness, diversity, and scenario coverage.
(2) Through extensive verification, \METHODNAME{} produces datasets that are \textbf{accurate and diverse while covering real-world scenarios}. Moreover, our approach demonstrates \textbf{strong generalizability to downstream tools}, underscoring its practical applicability.
(3) We fine-tune Qwen3-8B with training data generated by our pipeline and evaluate the \METHODNAME{}-8B on three widely used benchmarks, BFCL~\cite{BFCL}, API-Bank~\cite{DBLP:conf/emnlp/LiZ000YLHL23}, and ACEBench~\cite{DBLP:journals/corr/abs-2501-12851}.  \METHODNAME{}-8B consistently outperforms open-source LLMs of comparable scales and remains highly competitive with API-based models.

\section{Related Work}

\noindent \textbf{LLM Function-Calling Paradigms.}
Equipping LLMs with executable tools enables reliable and specialized problem solving~\cite{DBLP:conf/iclr/QinLYZYLLCTQZHT24}. Existing approaches mainly follow two paradigms: prompting and fine-tuning. Prompting leverages in-context tool specifications and demonstrations~\cite{DBLP:journals/tmlr/MialonDLNPRRSDC23, DBLP:journals/corr/abs-2308-00675}. ReAct~\cite{DBLP:conf/iclr/YaoZYDSN023} combines reasoning with API calls for multi-step tasks, but its performance is constrained by pretraining and degrades with increased tool complexity. These limitations motivate fine-tuning via supervised or reinforcement learning~\cite{DBLP:journals/corr/abs-2306-05301, DBLP:conf/emnlp/Abdelaziz0AKSPR24}. Representative methods include ToolACE~\cite{DBLP:conf/iclr/Liu0ZHYL0GLY0WN25}, which constructs large-scale tool-use corpora automatically, and RL-based approaches such as ToolRL~\cite{DBLP:journals/corr/abs-2504-13958} and AWPO~\cite{DBLP:journals/corr/abs-2512-19126}, which enhance function-calling through policy optimization with reasoning rewards.

\noindent \textbf{Data Synthesis for Function-Calling.}
As LLMs advance, reliance solely on human-authored corpora becomes insufficient for sustained progress~\cite{DBLP:journals/corr/abs-2401-02524}. To expand supervision without heavy annotation costs, recent works adopt prompt-driven transformations to augment existing datasets. For example, \citet{DBLP:conf/iclr/YuJSYLZKLWL24} proposes targeted prompting to elicit rare skills and long-tail behaviors from base models. However, purpose-driven data remain limited in tool-use scenarios. Prior efforts adapt resources from adjacent domains~\cite{DBLP:conf/acl/0002ACDCMAKMKL24} or synthesize samples around public APIs~\cite{DBLP:conf/nips/LiuHZZLKTYLFNYS24}. Moreover, ToolACE~\cite{DBLP:conf/iclr/Liu0ZHYL0GLY0WN25} constructs large-scale function-calling corpora via automated pipelines, while ToolForge~\cite{DBLP:journals/corr/abs-2512-16149} introduces automated tool-use synthesis with reduced reliance on real APIs.

Our work is distinguished from the most relevant studies~\cite{DBLP:conf/iclr/QinLYZYLLCTQZHT24, DBLP:conf/nips/LiuHZZLKTYLFNYS24, DBLP:conf/iclr/Liu0ZHYL0GLY0WN25} in the following aspects. Prior works often rely on annotated or synthetic APIs, which lack reliability and struggle to scale across larger tool sets. These approaches also face limitations in diversity, quality, and coverage. In contrast, \METHODNAME{} is built upon reliable tools drawn from public benchmarks and offers stronger scalability. Furthermore, by leveraging a multi-agent dialogue generation framework and a multi-stage verification system, \METHODNAME{} produces tool-use training data with richer diversity, higher quality, and broader scenario coverage than prior works.

\section{Preliminary}
Given a user query $q$ and a candidate tool set $\mathcal{T}=\{t_1,\ldots,t_n\}$, where each tool $t_i$ is defined by a name, a usage description, and a schema specifying required and optional parameters, the goal is to generate a valid sequence of tool calls by selecting tools and filling arguments with appropriate values and units.
This process can be framed as follows:
\begin{equation}
\mathcal{S}=\big[t_1(a_1),\ldots,t_k(a_k)\big]=g_\phi\!\big(q,\mathcal{T}\big),
\label{eq:tool-call-reform}
\end{equation}
where $g_\phi(\cdot)$ denotes an LLM with parameters $\phi$, $k$ is the number of invocations, and $a_i$ denotes the argument payload for the $i$-th call ($1\le i\le k$), which is a set of parameter-value pairs, that is,
$a_i=[\,r_1\!:\!w_1,\,r_2\!:\!w_2,\ldots,r_\ell\!:\!w_\ell\,]$,
with parameter names $r_j$ and corresponding values $w_j$.
The query $q$ may be a single turn or a full multi-turn history.

For fine-tuning, the pair $(q,\mathcal{T})$ is treated as input context, while the gold-standard tool-call sequence $\mathcal{S}$ serves as the supervised target. Formally, the training samples can be represented as $\{\,\langle q,\mathcal{T}\rangle,\mathcal{S}\,\}$, where each training instance connects a user query and the candidate tools with the corresponding sequence of tool invocations. 

\begin{figure*}[t]
  \centering  \includegraphics[width=1\textwidth]{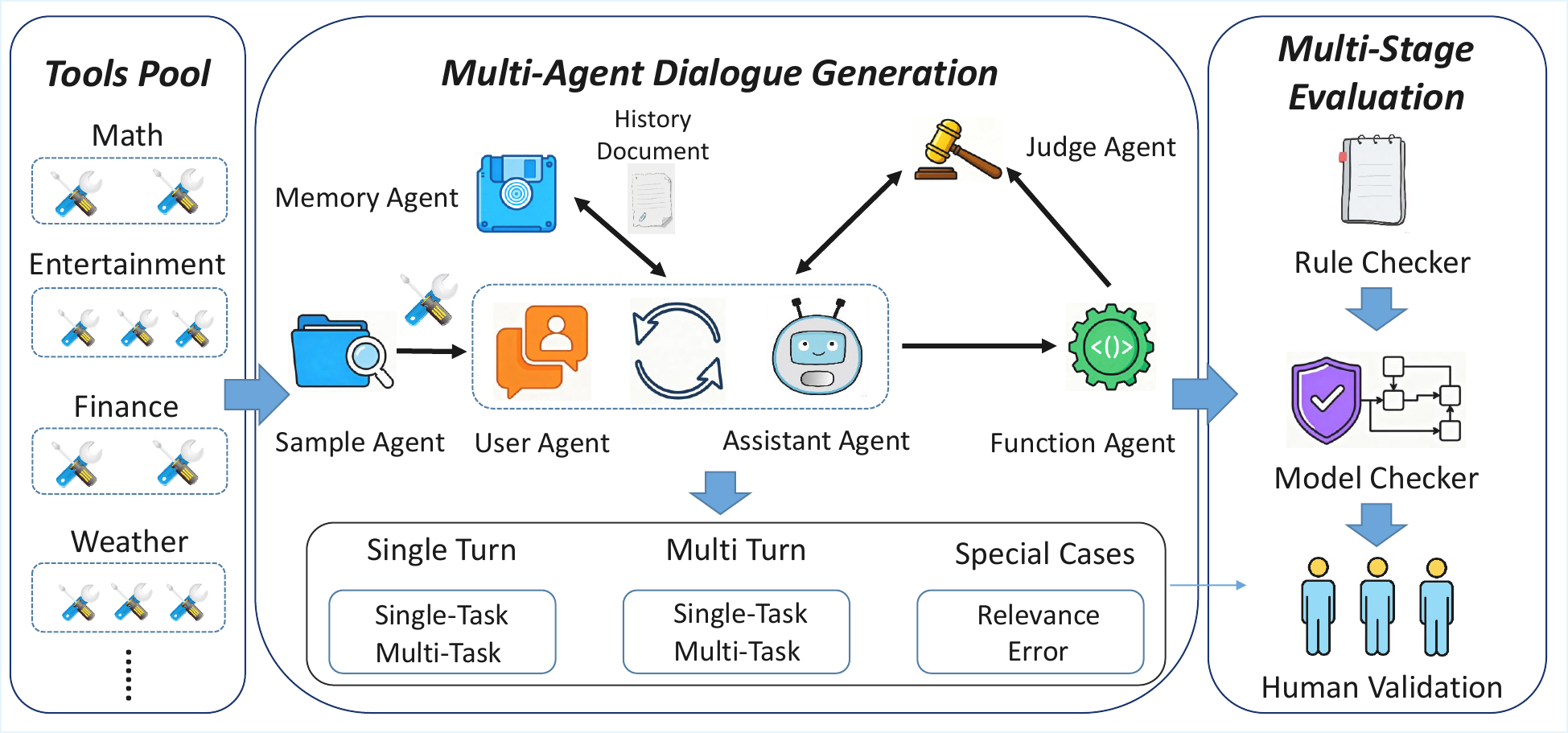}
  \vspace{-4mm}
  \caption{The overall framework of \textbf{\METHODNAME{}}, consisting of a \textbf{Tool Pool} built from reliable tools in open-source benchmarks, a \textbf{Multi-Agent Dialogue Generation} system, and a \textbf{Multi-Stage Evaluation} process.}
  \vspace{-3mm}
  \label{overall}
\end{figure*}

\section{Data Generation Pipeline}
Synthetic data plays a vital role in enhancing the function-calling capability of LLMs~\cite{DBLP:conf/iclr/Liu0ZHYL0GLY0WN25}. We propose an automated pipeline \METHODNAME{} that employs a structured, three-stage process to generate high-quality training data for tool-augmented LLMs, as illustrated in Figure~\ref{overall}.
First, reliable and functionally diverse tools are selected from public datasets and placed into the \textbf{Tool Pool}. 
Next, the \textbf{Multi-Agent Dialogue Generation} module leverages a multi-agent-assisted framework to produce tool-use dialogues that are diverse, accurate, and broadly representative. 
Finally, the \textbf{Multi-Stage Evaluation} module systematically examines the correctness of the generated dialogues to guarantee training data quality. 

\subsection{Tool Pool}
The quality of training data for enhancing LLM tool-use abilities depends critically on the reliability and coverage of the underlying APIs. However, existing datasets often rely on manually annotated APIs or handcrafted synthetic pipelines, which are either costly to scale or difficult to extend to real-world tools, limiting model generalizability.

To address these issues, \METHODNAME{} constructs a curated Tool Pool entirely sourced from the BFCL evaluation set~\cite{BFCL}. The pool is built through a two-stage filtering process: GPT-4o~\cite{DBLP:journals/corr/abs-2303-08774} is first used to perform semantic clustering over all tools to remove redundant or highly similar ones, followed by light human verification to ensure correctness and usability. This process yields a curated Tool Pool of 1,000 tools.
Since tools in BFCL are manually designed to cover a broad range of real-world usage scenarios, we further verify that the curated pool maintains high functional diversity across multiple domains. By relying on widely used and practically accessible tools, \textbf{the Tool Pool balances reliability, diversity, and scalability, and serves as a critical foundation for the subsequent data generation modules in \METHODNAME{}}.

\subsection{Multi-Agent Dialogue Generation}
For LLMs, the quality of fine-tuning data significantly affects downstream performance~\cite{DBLP:conf/naacl/LiuKLJL25}. 
We propose a multi-agent dialogue generation module that leverages a multi-agent framework to assist the dialogue generation system in synthesizing function-calling dialogues. By leveraging agent interaction and collaboration, this module produces dialogues that are high-quality, diverse, and broadly representative, which is crucial to enhance tool-use capabilities of LLMs.

\subsubsection{Multi-Agent Framework}
The multi-agent framework includes four LLM-based agents: \textit{Sample Agent}, \textit{Memory Agent}, \textit{Function Agent}, and \textit{Judge Agent}. We adopt Gemini-2.5 Pro~\cite{DBLP:journals/corr/abs-2507-06261} as the backbone of all agents, which is an engineering choice due to Gemini-2.5 Pro's \textit{stable API-following performance} rather than an algorithmic requirement. Importantly, the multi-agent framework is \textit{model-agnostic}; Appendix~\ref{app-prompt} shows that \textbf{replacing proprietary models with strong open-source alternatives retains the overall workflow and leads to only moderate data quality degradation and slight downstream performance drops}. System prompts and detailed agent descriptions and analyses are also in Appendix~\ref{app-prompt}.

\textbf{Sample Agent} selects a subset $\mathcal{T}_s \subseteq \mathcal{T}$ from the global Tool Pool, 
including target tools $\mathcal{T}_{\text{target}}$ and distractors $\mathcal{T}_{\text{dist}}$. 
Target tools are grouped by semantic or functional similarity, while distractors are chosen 
with varied relevance to increase realism and require the model to distinguish 
between strongly and weakly related tools. 

\textbf{Memory Agent} records dialogue instances 
$\mathcal{D}=\{d_1,\ldots,d_k\}$ and assigns each a type label 
$\tau(d_i)\in\mathcal{C}$ reflecting the semantic context of tool usage,
where the same math tool may address architectural geometry in one case 
and year calculations in another. 
It then summarizes past pairs $(d_i,\tau(d_i))$ and guides the generator 
to produce new dialogues with unseen $\tau$, improving semantic diversity.

\textbf{Function Agent} selects one or more tools 
$\{t_1, \ldots, t_k\} \subseteq \mathcal{T}_s$ to address the user query $q$, 
and extracts both required and optional parameters for each tool $t_i$. 
To enhance diversity and realism, a slot-selection strategy is applied: during parameter 
extraction, the agent randomly chooses a subset $\mathcal{P}' \subseteq \mathcal{P}$ 
of optional parameters to instantiate, resulting in varied and realistic tool calls.

\textbf{Judge Agent} ensures rigorous quality control by selecting the best dialogue $d^*$
from $N$ candidates $\{d_1,\ldots,d_N\}$ (default $N=4$) each generated round.
Evaluation considers \textbf{problem significance} and \textbf{tool appropriateness}, with \textbf{positional bias controlled} as discussed in Appendix~\ref{app-prompt}.
$N=4$ balances selection quality and generation efficiency.
The selected $d^*$ serves as the final output.

\subsubsection{Dialogue Generation System}
We design a dialogue generation system to produce three types of function-calling dialogues: \textbf{single-turn}, \textbf{multi-turn}, and \textbf{special-case}.
The first two types cover both single-task and multi-task settings, enabling the simulation of simple requests and more complex interaction flows. The special cases, on the other hand, are designed to cover scenarios for relevance checking and error detection.
The overall system is composed of two agents, a user agent $U$ and an assistant agent $A$, both instantiated with Gemini-2.5 Pro, that interact with each other to generate task-oriented conversations.

The detailed algorithm of the multi-agent dialogue generation module and dialogue examples are provided in Appendix~\ref{dia}.
The dialogue generation system grounds each conversation in the sampled tool set $\mathcal{T}_s$ drawn from the curated pool. 
Every dialogue contains user requests that can be handled by $\mathcal{T}_s$ and tool call answers that explicitly specify the selected tools and the full argument payload. 
The assistant's action space is $\mathcal{A}_{\text{asst}}=\{\texttt{call},\,\texttt{ask},\,\texttt{answer}\}$. 
The assistant can issue a tool call $t(a)$, request clarification, summarize tool outputs, or provide a direct non-tool reply when a call is unnecessary. 
The system records the chosen tool $t$ and the payload $a$ in the turn-level state and appends them to the \textit{turn-level trajectory} for later evaluation and reuse. 
These recorded trajectories, after being categorized by the memory agent, are leveraged to guide subsequent dialogue generation, preventing the system from reproducing similar contexts and encouraging scenario diversity.

At turn $t$, the user agent $U$ issues a request $q_t$ based on the tool set $\mathcal{T}_s$. The assistant agent $A$ receives the request $q_t$ together with the dialogue history $\mathcal{D}$, follows the system prompts, and generates the next action. In single-turn dialogues, $A$ constructs a problem solvable with $\mathcal{T}_s$, produces a tool call, and returns the final answer. In multi-turn dialogues, $A$ alternates with $U$ by requesting additional information when constraints are missing, and once the target length is reached or a stop signal triggers, the model outputs either \texttt{call} or \texttt{answer}. In special cases, prompts guide the generation of unsolved or erroneous samples, such as mismatched tools or invalid parameter values, to support relevance checking and error detection. 

\subsection{Multi-Stage Evaluation} 
We introduce a multi-stage evaluation system to assess the quality of synthesized dialogues, since inaccurate training data may significantly weaken the models' ability to understand and execute functions. This system integrates a rule-based checker and a model-based checker, with final verification by human experts to ensure the accuracy of the resulting training data. More details can be found in Appendix~\ref{app-rule}.

\noindent \textbf{(1) Rule Checker.}
Without executing tools, the Rule Checker performs basic compliance checks on synthesized dialogues to quickly filter out samples with format and alignment issues. 
It validates four aspects: completeness of the tool definition, compliance of the call format and parameters, soundness of the dialogue structure, and consistency between tool results and the assistant's statements (rule details in Appendix~\ref{app-rule})
Each rule returns a \textit{pass} or \textit{fail} flag with a hint for correction, and outputs are aggregated for the downstream Model Checker and Human Validation.

\noindent \textbf{(2) Model Checker.}
After format screening, the Model Checker employs GPT-4o~\cite{DBLP:journals/corr/abs-2303-08774} to evaluate semantic quality and task completion beyond the static rules (the prompt is detailed in Appendix~\ref{app-rule}). Unlike the Judge Agent that ranks candidate dialogues \textit{during} generation, the Model Checker works \textit{post-hoc} to verify finalized dialogues. Given the dialogue context and tool calls, it checks faithfulness, task satisfaction, and compliance, returning a rationale and a confidence score. Only samples with confidence scores above the threshold ($\theta = 0.75$) are retained.

\noindent \textbf{(3) Human Validation.}
After the Rule Checker and the Model Checker complete automated screening, the remaining error rate is \textbf{below 5\%}. Manual analysis shows that \textbf{over 80\%} of the errors after automated screening stem from parameter extraction rather than function selection or intent understanding. Samples that fail the Rule Checker or obtain low-confidence from the Model Checker are routed to a human review, with higher priority given to errors that prevent correct function execution and errors in high-impact functions that are commonly used in core real-world scenarios.
These samples account for about \textbf{5\%} of all samples. Human experts review each of these samples and provide revisions; the approved or revised dialogues are added to the training data after passing the second-pass human validation. Overall, the Human Validation stage only costs \textbf{$\sim$15 human-hours}.  

\noindent \textbf{Overall comparison with other strong synthetic function-calling datasets.}
Appendix~\ref{data} provides detailed analyses of quantity, quality, coverage, diversity of the final training data. Compared to other strong synthetic function-calling datasets, our dataset achieves higher quality through richer multi-tool and multi-turn compositions, broader coverage across diverse scenarios, and balanced parameter utilization patterns, while maintaining efficient construction by reusing existing tool schemas and avoiding additional manual designs.
\begin{table*}[h]
\centering
\caption{\textbf{Accuracy} on the \textbf{In-Domain} BFCL dataset. \METHODNAME{}-8B is fine-tuned on Qwen3-8B using training data generated by our pipeline. The best results in each category are in \textbf{bold} and the second best results are \underline{underlined}. We report mean and standard deviation ($\pm$SD) of the \textit{Overall} accuracy based on three independent runs.}
\renewcommand{\arraystretch}{1.25}
\resizebox{\textwidth}{!}{
\begin{tabular}{lcccccccccc}
\toprule
 & \multicolumn{5}{c}{\textbf{Non-Live}} & \multicolumn{5}{c}{\textbf{Live}} \\
\cmidrule(lr){2-6} \cmidrule(lr){7-11}
\textbf{Models} & \textit{Simple} & \textit{Multiple} & \textit{Parallel} & \textit{Parallel} & \textit{Overall} &\textit{Simple} & \textit{Multiple} & \textit{Parallel} & \textit{Parallel} & \textit{Overall}\\
 &  &  &  & \textit{Multiple} &  &  &  & & \textit{Multiple} \\
\midrule
\textbf{GPT-4o} & 76.50 & 91.00 & 90.00 & 78.00 & 83.88 & 70.54 & 70.75 & 62.50 & 66.67 & 70.54 \\
\textbf{GPT-4o-mini} & 80.33 & 92.00 & 89.50 & 90.50 & 88.08 & 79.46 & 76.26 & \underline{87.50} & 70.83 & 76.91\\
\textbf{Gemini-2.5-Pro} & 78.67 & 94.00 & 93.50 & 92.00 & 89.54 & 85.66 & 74.36 & \underline{87.50} & 83.33 & 76.83 \\
\midrule
\textbf{LLaMA-3.1-8B-Instruct} & 71.00 & 95.00 & 87.50 & 82.50 & 84.00 & 72.87 & 71.13 & 50.00 & 45.83 & 59.96\\
\textbf{LLaMA-3.1-70B-Instruct} & 78.33 & \underline{97.00} & \underline{95.50} & \textbf{94.00} & \underline{91.21} & 83.33 & 75.59 & \underline{87.50} & 58.33 & 76.91\\
\textbf{xLAM-8B} & 73.83 & 93.50 & 87.50 & 83.50 & 84.58 & 75.58 & 66.57 & 56.25 & 54.17 & 67.95\\
\textbf{ToolACE-8B} & 81.17 & 96.00 & 94.00 & \underline{93.00} & 91.04 & 82.95 & 79.58 & 75.25 & \textbf{85.12} & \underline{80.73}\\
\textbf{DeepSeek-V3} & 76.58 & 94.50 & 92.00 & 92.00 & 87.77 & \underline{86.05} & 78.82 & 75.00 & 66.67 & 79.94\\
\textbf{Qwen3-8B} & 78.92 & 95.00 & 91.50 & 89.00 & 88.60 & 80.23 & 77.21 & 81.25 & 75.00 & 77.79 \\
\textbf{Qwen3-32B} & 79.58 & 95.00 & 92.00 & \underline{93.00} & 89.90 & 84.50 & \underline{80.44} & \textbf{93.75} & 66.67 & 81.13 \\
\textbf{Hammer2.1-7B} & 77.17 & 96.00 & 93.00 & 87.50 & 88.42 & 77.13 & 77.59 & \underline{87.50} & 70.83 & 77.50 \\
\textbf{Qwen-ToolRL-8B} & \underline{83.50} & 89.50 & 91.25 & 91.85 & 89.03 & 81.48 & 77.82 & 72.75 & 81.50 & 78.39 \\
\midrule
\textbf{\METHODNAME{}-8B (Ours)} & \textbf{86.85} & \textbf{97.25} & \textbf{96.15} & \underline{93.00} & \textbf{93.31 $\pm$ 0.42} & \textbf{88.25} & \textbf{80.50} & 81.50 & \underline{84.88} & \textbf{83.78 $\pm$ 0.37}\\
\bottomrule
\end{tabular}
}
\label{bfcl}
\end{table*}

\vspace{-2mm}
\section{Experiments}
\vspace{-2mm}
\subsection{Experimental Setup}
We fine-tune LLMs on the training data produced by our pipeline and evaluate the resulting models in a broad range of settings. Unless otherwise noted, we conduct SFT~\cite{DBLP:conf/iclr/HuSWALWWC22} on Qwen3-8B~\cite{DBLP:journals/corr/abs-2505-09388} using our training data and denote the resulting model by \METHODNAME{}-8B.
We compare \METHODNAME{}-8B against top-tier API-based and open-source foundation LLMs, as well as representative function-calling models trained with specialized function-calling training data including ToolACE-8B (fine-tuning LLaMA3.1-8B-Instruct on ToolACE-generated data), Hammer2.1-7B (based on Qwen 2.5 coder series), and Qwen-ToolRL-8B (fine-tuning Qwen3-8B with ToolRL datasets). Evaluations are conducted on three commonly used benchmarks: BFCL~\cite{BFCL}, API-Bank~\cite{DBLP:conf/emnlp/LiZ000YLHL23}, and ACEBench~\cite{DBLP:journals/corr/abs-2501-12851}, with all results averaged over three independent runs. More experimental details, including evaluation metric definitions and settings, are in Appendix~\ref{app-exp}.

\begin{table*}[htbp]
\centering
\caption{\textbf{Accuracy} on the \textbf{Out-of-Domain} API-Bank and ACEBench datasets. The best results in each category are in \textbf{bold} and the second best results are \underline{underlined}. We report mean and standard deviation ($\pm$SD) of the \textit{Overall} accuracy from our model based on three independent runs.
}
\renewcommand{\arraystretch}{1.25}
\resizebox{\textwidth}{!}{
\begin{tabular}{lccccccccccccc}
\toprule
 & \multicolumn{4}{c}{\textbf{API-Bank}} & & \multicolumn{7}{c}{\textbf{ACEBench}} \\
\cmidrule(lr){2-5} \cmidrule(lr){7-13}
 & & & & & & \multicolumn{6}{c}{\textbf{Normal}} & \multicolumn{1}{c}{\textbf{Special}} \\
\cmidrule(lr){7-12} \cmidrule(lr){13-13}
\textbf{Models} & Level-1 & Level-2 & Level-3 & Overall & & Atom & Single-Turn & Multi-Turn & Similar API & Preference & Overall & Overall \\
\midrule
\textbf{GPT-4o} & \underline{76.19} & 42.96 & 35.21 & 51.45 & & \textbf{90.00} & \textbf{78.00} & \textbf{77.00} & \textbf{85.00} & \textbf{83.00} & \textbf{82.60} & \textbf{87.60} \\
\textbf{GPT-4o-mini} & 74.69 & 45.93 & 40.77 & 53.80 & & 84.33 & 73.50 & \underline{66.50} & 77.00 & \underline{78.00} & 75.87 & 79.90 \\
\textbf{Gemini-1.5-Pro} & 75.43 & 43.26 & 41.51 & 53.40 & & 84.50 & 76.80 & 64.50 & \underline{80.00} & \underline{78.00} & \underline{76.76} & 79.00 \\
\midrule
\textbf{Qwen3-8B} & 71.68 & 52.24 & 42.75 & 55.56 & & 80.00 & 65.50 & 51.00 & 68.00 & 60.00 & 64.90 & 76.67 \\
\textbf{Qwen-ToolRL-8B} & 73.07 & \underline{60.81} & \underline{47.21} & \underline{60.36} & & 79.00 & 68.50 & 52.00 & 66.00 & 60.00 & 65.10 & 78.67 \\
\textbf{LLaMA-3.1-8B-Instruct} & 71.18 & 37.04 & 35.88 & 48.03 & & 51.90 & 39.80 & 28.00 & 66.00 & 46.00 & 46.34 & 46.60 \\
\textbf{ToolACE-8B} & 75.94 & 47.41 & 45.27 & 56.21 & & 84.00 & 74.50 & 61.00 & 74.00 & 58.00 & 70.30 & 1.00 \\
\midrule
\textbf{\METHODNAME{}-8B (Ours)} & \textbf{79.17} & \textbf{64.09} & \textbf{51.11} & \textbf{64.79 $\pm$ 0.41} & & \underline{88.00} & \underline{77.00} & 65.00 & 74.00 & 64.00 & 73.60 $\pm$ 0.32 & \underline{83.67 $\pm$ 0.35}  \\
\bottomrule
\end{tabular}
}
\label{ood}
\end{table*}

\subsection{Main Results}
To comprehensively assess function-calling performance, we compare \METHODNAME{}-8B with baselines on both in-domain and out-of-domain datasets. 
The in-domain evaluation measures performance on data aligned with the training distribution, while the out-of-domain evaluation assesses its generalizability to unseen scenarios.

\textbf{In-Domain Evaluation.}
Since the tools in pool are drawn directly from BFCL, we treat BFCL as the in-domain benchmark. As shown in Table~\ref{bfcl}, \METHODNAME{}-8B achieves 93.31 on the Non-Live split and 83.78 on the Live split, showing strong and robust performance, and outperforming the API-based models. \METHODNAME{}-8B achieves substantial gains over open-source models of similar scale and on some subsets matches or outperforms much larger models such as Qwen3-32B. Compared with the prior open-source function-calling SOTA ToolACE-8B, our model improves overall accuracy by \textbf{2.5\% relative} on Non-Live and \textbf{3.8\% relative} on Live. 
Overall, these results indicate that, \textbf{for function-calling, strong and systematic alignment between training data distribution and real tool semantics can significantly narrow the performance gap between smaller models and much larger ones}.

\vspace{-0.7mm}

\textbf{Out-of-Domain Evaluation.}
To evaluate the generalizability of our fine-tuned model, we conduct experiments on two out-of-domain benchmarks: API-Bank and ACEBench. As shown in Table~\ref{ood}, API-based models maintain a clear advantage over open-source ones, with GPT-4o reaching 85.10 on ACEBench. Open-source models fine-tuned for function calling achieve competitive performance but fall short. Compared with the prior open-source SOTA, our model achieves 64.79 on API-Bank and 78.64 (mean of Normal Overall and Special Overall) on ACEBench, yielding 7.3\% and 9.4\% relative gains. Notably, \METHODNAME{}-8B performs on par with GPT-4o-mini and Gemini-1.5-Pro. These results validate that \textbf{the training data generated by our pipeline effectively enables model generalization to unseen scenarios}.

\vspace{-0.7mm}

\textbf{General Abilities.} We also find that \textbf{fine-tuning on the training data generated by our pipeline preserves the model’s general abilities while substantially improving function-calling ability} (more details in Appendix~\ref{gen}).

\begin{figure}[t]
  \subfigure[Non-Live]{
  \includegraphics[width=0.23\textwidth]{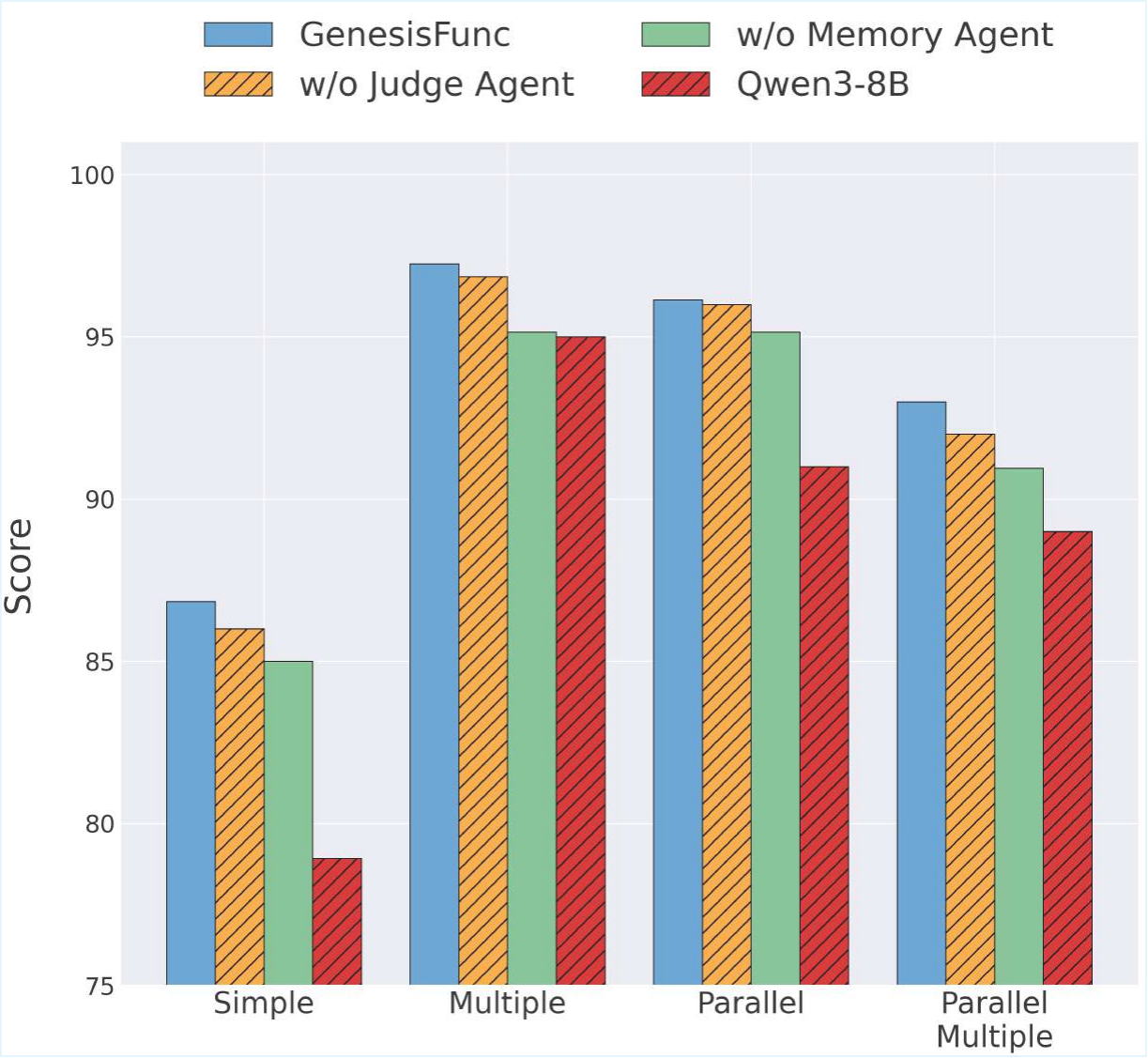}}
  \subfigure[Live]{
  \includegraphics[width=0.23\textwidth]{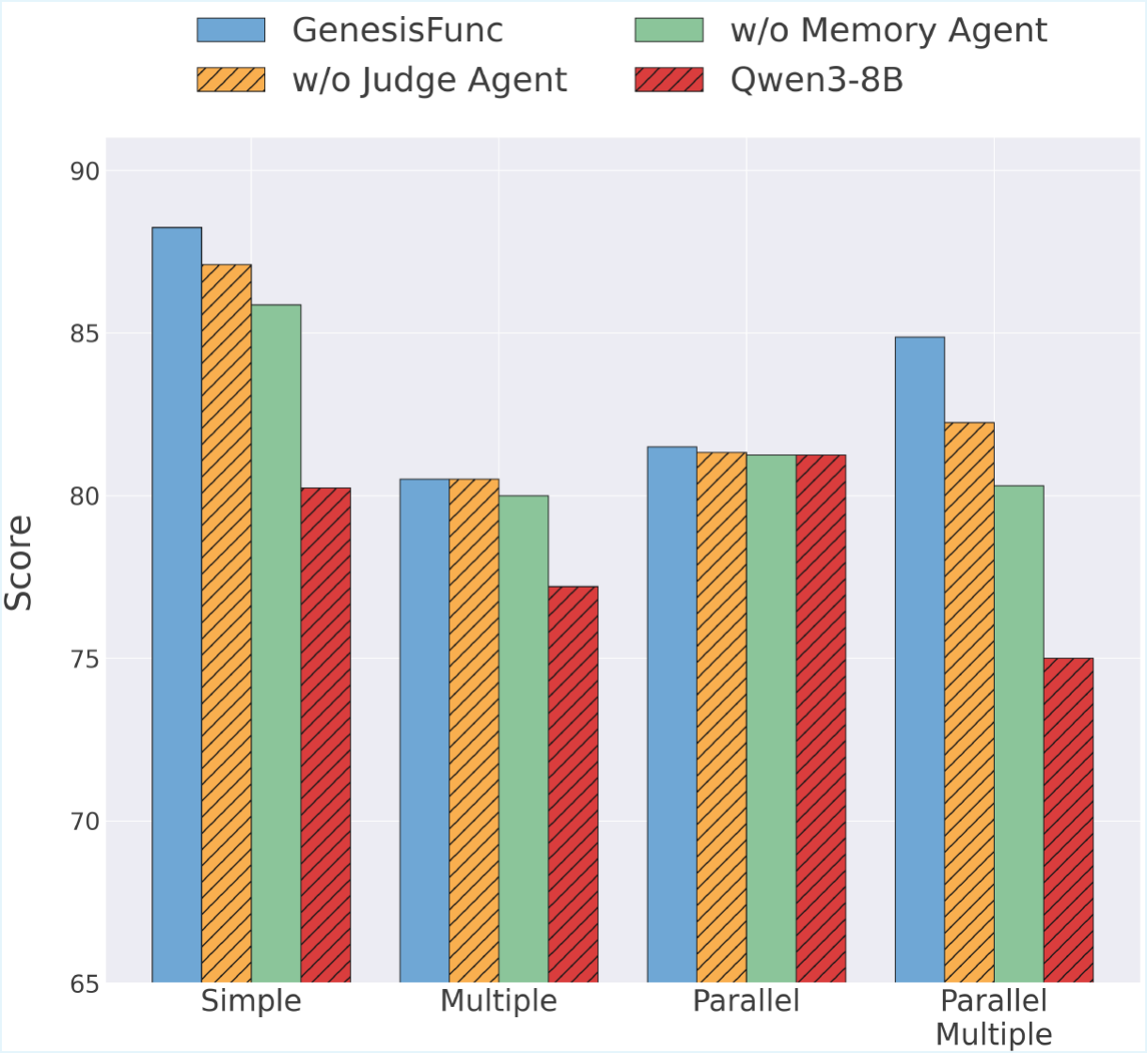}}
\vspace{-5mm}
\caption{Ablation study of the proposed agents. We separately remove the Judge Agent and the Memory Agent, and evaluate our \METHODNAME{}-8B on BFCL in (a) Non-Live and (b) Live settings.}
\label{abla-module}
\vspace{-5mm}
\end{figure}

\begin{figure}[t]
  \subfigure[Non-Live]{
  \includegraphics[width=0.23\textwidth]{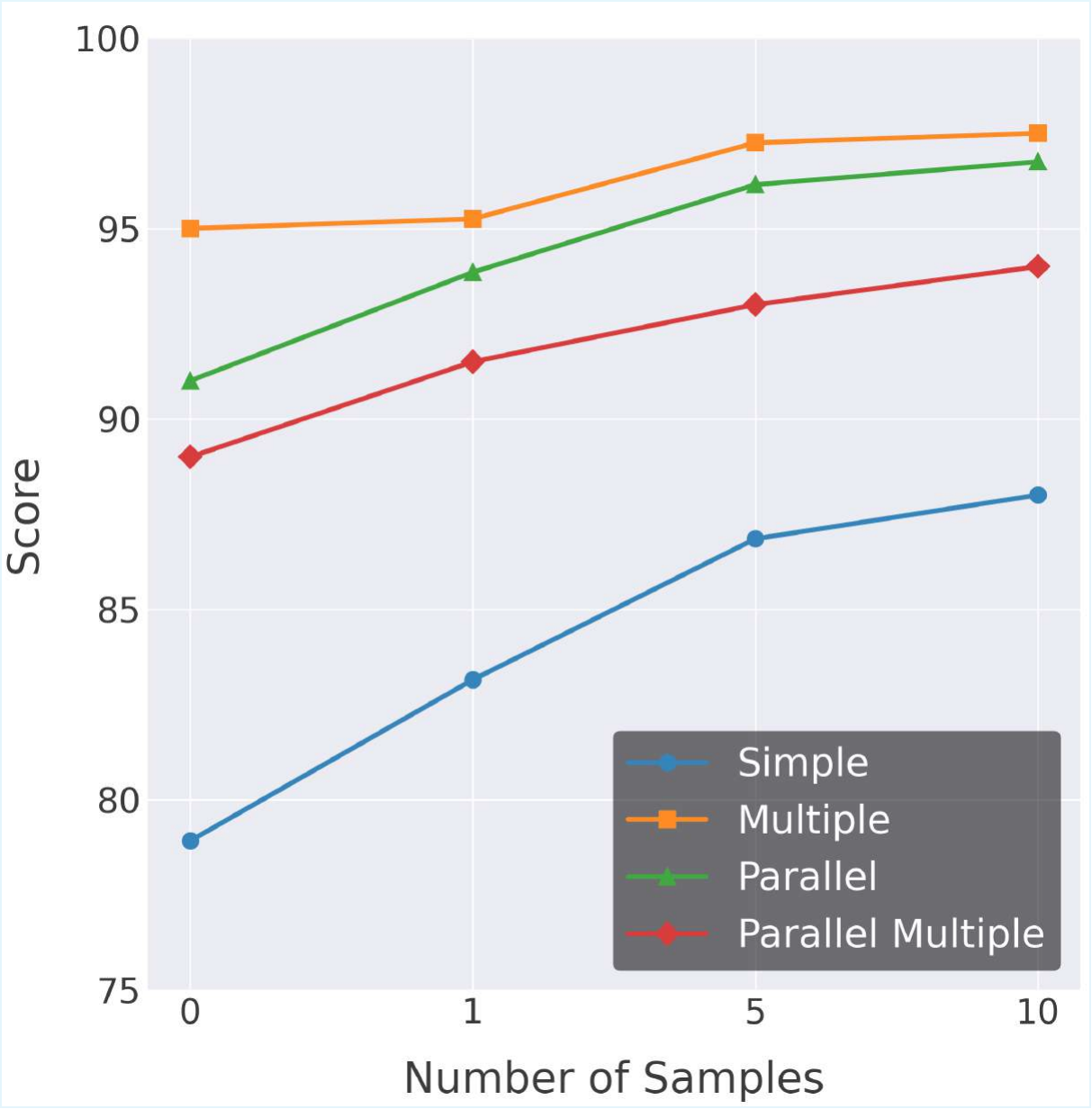}}
  \subfigure[Live]{
  \includegraphics[width=0.23\textwidth]{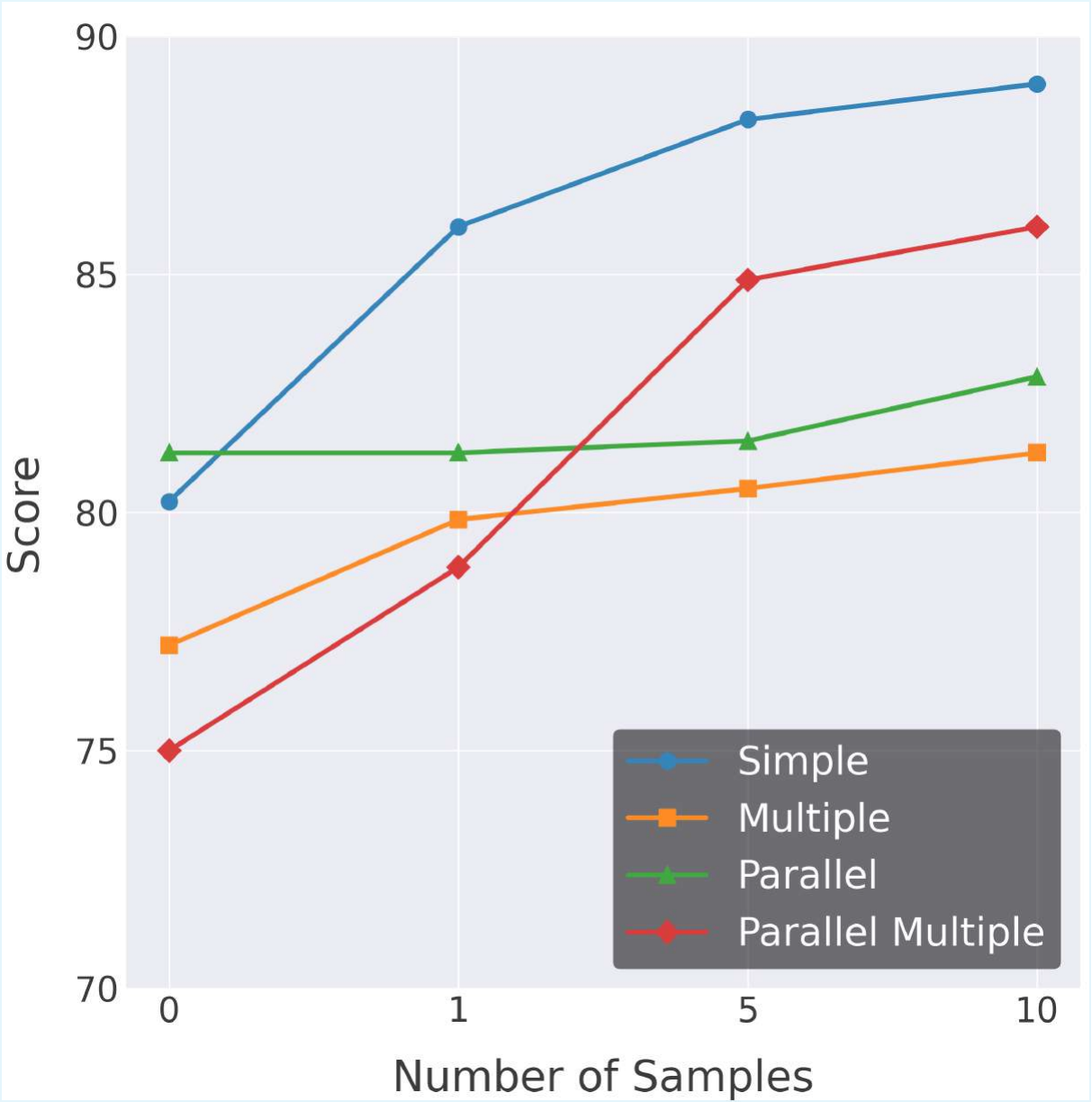}}
\vspace{-2mm}
\caption{Ablation study on the number of samples. We control the number of dialogues generated per run and evaluate on BFCL in (a) Non-Live and (b) Live settings.}
\label{abla-number}
\vspace{-5mm}
\end{figure}

\vspace{-3mm}
\subsection{Ablation Study}

\textbf{Ablation on Multi-Agent Framework.}
Our multi-agent framework improves function-calling ability through agent collaboration, where the Memory Agent enhances dialogue diversity and the Judge Agent ensures accuracy. We conduct ablation studies by removing either agent and fine-tuning Qwen3-8B with LoRA, noting that the Sample Agent and Function Agent are indispensable to the workflow. As shown in Figure~\ref{abla-module}, BFCL results indicate that removing the Judge Agent causes a clear performance drop, while removing the Memory Agent leads to significant degradation, highlighting the importance of accuracy and diversity and validating the effectiveness of both agents. We further compare alternative slot selection strategies in the Function Agent in Appendix~\ref{1}.

\begin{table*}[h]
\centering
\caption{\textbf{Accuracy} on ACEBench using \textbf{reinforcement learning}. The best overall results in each category are marked in bold. The second best results are \underline{underlined}.}
\label{rl-multi}
\renewcommand{\arraystretch}{1.25}
\resizebox{\textwidth}{!}{
\begin{tabular}{lccccccc}
\toprule
 & \multicolumn{6}{c}{\textbf{Normal}} & \multicolumn{1}{c}{\textbf{Special}} \\
\cmidrule(lr){2-7} \cmidrule(lr){8-8}
\textbf{Models} & Atom & Single-Turn & Multi-Turn & Similar API & Preference & Overall & Overall \\
\midrule
\textbf{Qwen3-8B} & 80.00 & 65.50 & 51.00 & 68.00 & 60.00 & 64.90 & 76.67 \\
\textbf{\METHODNAME{}-8B} & 88.00 & 77.00 & 65.00 & 74.00 & 64.00 & \underline{73.60} & \textbf{83.67} \\
\midrule
\textbf{\METHODNAME{}-8B-RL(all)} & 87.00 & 78.00 & 66.00 & 72.00 & 62.00 & 73.00 & 82.45 \\
\textbf{\METHODNAME{}-8B-RL(part)} & 88.00 & 79.00 & 70.00 & 75.00 & 64.00 & \textbf{75.20} & \underline{82.88} \\
\bottomrule
\end{tabular}
}
\end{table*}

\textbf{Impact of the Number of Samples.}
To assess how the number of generated samples influences function-calling ability, we use the same set of tools and only vary how many dialogues are produced for each tool.
Besides the default setting of 5 dialogues per tool, we also construct datasets with 1 and 10 dialogues per tool. We fine-tune Qwen3-8B on these datasets and report BFCL results in Figure~\ref{abla-number}. Overall accuracy steadily consistently rises with more samples per tool, with a substantial gain from 1 to 5 samples but only modest improvement from 5 to 10, indicating eventually diminishing returns once sufficient scenario diversity is reached.

\textbf{Ablation on Multi-Stage Evaluation.} Detailed results are in Appendix~\ref{ABA}. We find that models trained on data with the multi-stage evaluation module achieve higher accuracy across all conditions than those trained on non-evaluated data, verifying the effectiveness of this module.

\begin{table}
\renewcommand{\arraystretch}{0.9}
\centering
\caption{\textbf{Overall Accuracy} on BFCL, API-Bank and ACEBench adding the tools in ACEBench. The best overall results in each category are marked in bold. The second best results are \underline{underlined}.}
\resizebox{0.45\textwidth}{!}{
\begin{tabular}{lccc}
\toprule
\textbf{Models} & \textbf{BFCL} & \textbf{API-Bank} & \textbf{ACEBench}\\
\midrule
\textbf{Qwen3-8B}   & 83.20  & 55.56  & 70.79\\
\textbf{ToolACE-8B}   & 85.89  & 56.21  & 35.65\\
\midrule
\textbf{\METHODNAME{}}   & \textbf{88.55}  & \underline{64.79}  & \underline{78.64}\\
\quad + \textbf{ACEBench}   & \underline{87.89} & \textbf{65.11} & \textbf{81.87}\\
\bottomrule
\end{tabular}
}
\label{scal}
\end{table}

\vspace{-0.7mm}

\subsection{Scalability and Reinforcement Learning}

\textbf{Scalable to More Tools.} To assess scalability of our pipeline to more downstream tools, we add tools defined in ACEBench into the pool and use our pipeline to generate high-quality training data for fine-tuning. As shown in Table~\ref{scal}, performance on ACEBench improves markedly from 78.64 to 81.87 after adding these tools, due to better tool alignment. Meanwhile, aggregate results on BFCL remain comparable at 87.89 versus 88.55, and on API-Bank at 65.11 versus 64.79, indicating that introducing additional tool definitions does not notably degrade performance on other benchmarks. 
Overall, \textbf{our pipeline scales well to broader tool inventories and yields consistent gains on targeted benchmarks with no observable degradations on other datasets, across different model sizes and different backbone architectures}, as shown in Appendix~\ref{scale} and Appendix~\ref{backbone}.

\textbf{Enhancing Function-Calling Ability in Multi-Turn Dialogues via RL.}
Inspired by prior works~\cite{DBLP:journals/corr/abs-2504-13958, DBLP:journals/corr/abs-2505-17667}, one research question is \textit{whether applying reinforcement learning (RL) on our data can further enhance the model’s function-calling capability, particularly in multi-turn dialogue scenarios where performance is less satisfactory}.
We conduct two sets of experiments: \METHODNAME{}-8B-RL(all) applies RL instead of SFT to Qwen3-8B using our training data; \METHODNAME{}-8B-RL(part) focuses on data in multi-turn scenarios, where the model is first SFT-trained on single-turn and special-case data and then RL-trained with multi-turn dialogues.
To encourage deeper reasoning during training, we enable the built-in ``thinking mode'' of Qwen3-8B and augment the training samples with explicit reasoning traces. We use GRPO~\cite{DBLP:journals/corr/abs-2402-03300} for RL, with rewards designed around two dimensions: format compliance and functional correctness. More details are in Appendix~\ref{RL}.

Table~\ref{rl-multi} compares the performance of different training strategies on ACEBench. The model SFT-ed on our data achieves 83.67. Compared with the default SFT setup, 
\METHODNAME{}-8B-RL (part) improves Normal tasks substantially from 73.60 to 75.20, with multi-turn performance improved from 65.00 to 70.00, while maintaining 82.88 on Special subset, confirming that \textbf{targeted RL training enhances long-context reasoning and handling complex interaction, improving generalization on multi-turn tasks while maintaining stability on Special cases}. These results highlight the \textbf{advantage of combining SFT and RL} to improve function-calling abilities of the model. 

Notably, both SFT and RL incur low training costs, as summarized in Appendix~\ref{CO}.
\vspace{-0.7mm}

\section{Conclusion}
This paper introduces \METHODNAME{}, an automated data-generation pipeline for strengthening the function-calling abilities of LLMs. Beginning with a reliable tool set curated from open-source benchmarks, \METHODNAME{} leverages a coordinated multi-agent framework that assists a dialogue generator in producing high-quality, diverse, and representative function-calling training data, while remaining model-agnostic in design. In extensive experiments, models trained with \METHODNAME{} achieve state-of-the-art performance, marking a concrete advance in tool-augmented AI agents, and the methodology can be extended to more agentic and complex function-calling tasks using proprietary or open-source LLMs in future work.

\section*{Limitations}
Despite the effectiveness of \METHODNAME{}, several important limitations still remain.
First, \METHODNAME{}-8B achieves competitive tool-use performance but still falls short of API-based models like GPT-4 in broader reasoning and comprehension. Enhancing general abilities alongside function-calling remains an open challenge.
Second, our training data does not yet fully encompass highly complex multi-turn scenarios that require tightly coupled tool sequences. 
In future work, we plan to focus more on challenging settings, such as agentic workflows and complex benchmarks. Nevertheless, we firmly believe that, in principle, these more demanding scenarios can also be addressed by extending the methodology developed in this work.

\section*{Ethical Considerations}

\paragraph{AI Assistant Use} We used LLMs to assist with improving grammar, clarity, and wording in parts of this work. The use of LLMs was limited to language refinement, with all ideas, analyses, and conclusions solely developed by the authors.


\clearpage
\newpage
\appendix
\section{Details of \METHODNAME{}}

\subsection{Multi-Agent Framework}\label{app-prompt}

\begin{figure*}[ht]
    \centering
    \includegraphics[width=1\linewidth]{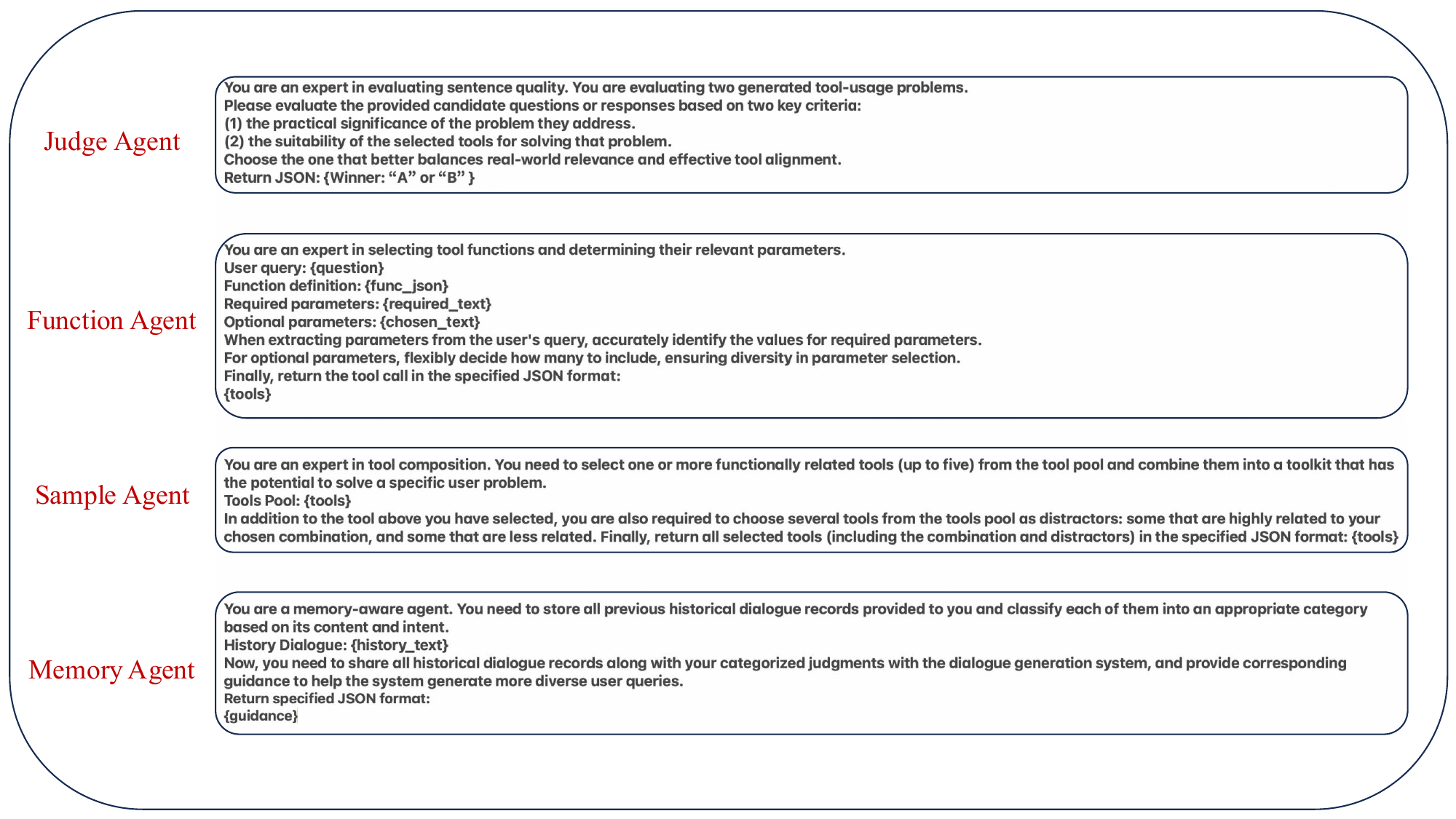}
    \caption{The system prompts of each agent in multi-agent framework.
}
    \label{system}
\end{figure*}

\paragraph{System Prompts in Multi-Agent Framework}
In our proposed multi-agent framework, we employ four distinct agents, Sample Agent, Memory Agent, Function Agent, and Judge Agent, to assist dialogue generation by producing diverse and high-quality conversations. Figure~\ref{system} presents the complete system-level prompts that define the roles and behaviors of the Sample Agent, Memory Agent, Function Agent, and Judge Agent. These prompts are used as system prompts throughout the generation process, with bracketed segments serving as dynamically filled placeholders.

\paragraph{Details of Sample Agent}
To construct a sampled subset $\mathcal{T}_s \subseteq \mathcal{T}$ from the global \textbf{Tools Pool}, 
the sample agent employs GPT-4o~\cite{DBLP:journals/corr/abs-2303-08774} to evaluate the semantic and functional similarity between tools. 
Given two tool descriptions that specify their name, purpose, and parameter schema, 
the model assigns a similarity score $r \in [0,1]$, where $r=1$ indicates nearly identical functionality and $r=0$ indicates entirely unrelated purposes. 
Each tool pair is scored with a fixed prompt and temperature~$=0$ to ensure deterministic outputs. 
Tools with $r>0.75$ or sharing similar functions are grouped as target tools $\mathcal{T}_{\text{target}} = \{t_1,\ldots,t_m\}$. 
Distractors $\mathcal{T}_{\text{dist}} = \{t_{m+1},\ldots,t_{m+n}\}$ are sampled from other clusters according to their relevance scores, 
categorized as high ($r>0.6$), medium ($0.3<r\le0.6$), and low ($r\le0.3$).  
All threshold values and sampling ratios were empirically tuned for consistency and reproducibility.

\paragraph{Details of the Judge Agent}
The Judge Agent selects the best dialogue from multiple candidates based on problem significance and tool appropriateness. Prior literature has documented that LLM-based evaluators can exhibit positional bias in pairwise comparisons, where judgments may systematically favor specific positions within prompts~\cite{DBLP:conf/emnlp/LiW0W0G024}.
To mitigate this, candidate dialogues are randomly shuffled before being evaluated by the Judge Agent. Additionally, we perform an A/B swap test on a subset of pairs, where dialogues are evaluated under swapped ordering. We observe no consistent directional preference across swapped conditions, indicating that positional bias is unlikely to be a major confounding factor in our selection process. This analysis enhances the reliability of the Judge Agent’s decisions and helps ensure that the data pipeline’s improvements are not driven by spurious positional effects.

\paragraph{Reproducibility}
Our Multi-Agent Framework is model-agnostic and does not rely on any proprietary model capabilities or a specific combination of models. In practice, different stages of the pipeline impose different requirements on model behavior. In particular, tool-calling generation benefits from models with stable API-following performance, while other stages primarily require output consistency rather than advanced reasoning ability. 
In our implementation, Gemini-2.5 Pro~\cite{DBLP:journals/corr/abs-2507-06261} is adopted for tool-calling related agents due to its stable behavior, which reflects an engineering choice rather than an algorithmic requirement.
Moreover, replacing Gemini-2.5 Pro with open-source models such as Qwen3-32B~\cite{DBLP:journals/corr/abs-2505-09388} preserves the overall workflow and produces coherent training data, although data quality and downstream tool-calling performance are moderately reduced. Specifically, on BFCL, the performance decreases by about 0.5\% on the non-live split and 0.9\% on the live split. These results indicate that stronger models primarily affect data quality, while the pipeline logic and functionality remain unchanged, supporting both reproducibility and practical accessibility.

\begin{algorithm*}[t]
\caption{Dialogue Generation with Multi-Agent Coordination}
\label{alg}
\KwIn{Global tools pool $\mathcal{T}$; history $\mathcal{D}$ with type labels $\tau(\cdot)\!\in\!\mathcal{C}$; max turns $T_{\max}$; candidate count $N$}
\KwOut{Final dialogue \texttt{Dialog}; turn-level trajectory \texttt{traj}}

$\mathcal{T}_s \leftarrow \textsc{SampleAgent.Select}(\mathcal{T})$ \tcp*{targets + distractors}
$\texttt{summary},\,\texttt{forbidden} \leftarrow \textsc{MemoryAgent.Summarize}(\mathcal{D})$ \tcp*{avoid seen types}

\texttt{Dialog} $\leftarrow [\,]$;\quad \texttt{traj} $\leftarrow [\,]$;\quad
$s_0 \leftarrow \textsc{InitState}(\mathcal{T}_s, \texttt{summary})$

\For{$t \leftarrow 0$ \KwTo $T_{\max}$}{
  $\mathcal{CAND} \leftarrow \varnothing$ \tcp*{N candidates per round}

  \For{$i \leftarrow 1$ \KwTo $N$}{
    $q_t \leftarrow U.\textsc{Issue}(\mathcal{T}_s, s_t, \texttt{summary}, \texttt{forbidden})$ \tcp*{user request}
    $a_t \leftarrow A.\textsc{Plan}(q_t, s_t)$ \tcp*{$\mathcal{A}_{\text{asst}}=\{\texttt{ask},\texttt{call},\texttt{answer}\}$}

    \uIf{$a_t=\texttt{ask}$}{
      $o_t \leftarrow U.\textsc{Clarify}(a_t)$ \tcp*{supply missing constraints}
    }
    \uElseIf{$a_t=\texttt{call}$}{
      $\{t_1,\ldots,t_m\} \leftarrow \textsc{FunctionAgent.Select}(\mathcal{T}_s, q_t)$ \tcp*{one or more tools}
      $\mathcal{P}' \leftarrow \textsc{FunctionAgent.SlotSelect}(q_t)$ \tcp*{optional param subset}
      $\texttt{args} \leftarrow \textsc{FunctionAgent.Fill}(q_t,\mathcal{A}')$ \tcp*{instantiate required + optional}
      $r_t \leftarrow \textsc{FunctionAgent.Exec}(\{t_j\}, \texttt{args})$ \tcp*{simulate}
      $o_t \leftarrow A.\textsc{Summarize}(r_t)$ \tcp*{explicit tool outputs}
    }
    \Else{
      $o_t \leftarrow A.\textsc{DirectAnswer}(q_t)$ \tcp*{non-tool reply}
    }

    $\texttt{cand}_i \leftarrow (q_t, a_t, o_t)$;\quad $\mathcal{CAND} \leftarrow \mathcal{CAND} \cup \{\texttt{cand}_i\}$ \;
  }

  $\texttt{cand}^* \leftarrow \textsc{JudgeAgent.Select}(\mathcal{CAND};\,\texttt{significance},\,\texttt{suitability})$ \tcp*{pick best of $N$}
  \texttt{Dialog} $\leftarrow$ \texttt{Dialog} $\mathbin{\|}$ \texttt{cand}$^*$;\quad
  \texttt{traj} $\leftarrow$ \texttt{traj} $\mathbin{\|}$ \texttt{cand}$^*$ \;

  $\tau_{\text{new}} \leftarrow \textsc{InferType}(\texttt{cand}^*)$;\quad
  $\mathcal{D} \leftarrow \mathcal{D} \cup \{(\texttt{cand}^*, \tau_{\text{new}})\}$;\quad
  $\texttt{forbidden} \leftarrow \texttt{forbidden} \cup \{\tau_{\text{new}}\}$ \tcp*{diversity control}

  $s_{t+1} \leftarrow \textsc{UpdateState}(s_t, \texttt{cand}^*)$ \tcp*{record $t$, $a$, and payload $a$}

  \If{\textsc{Stop}$(s_{t+1})$}{\textbf{break}} \tcp*{single-turn met / constraints satisfied / max turns}
}
\Return{\texttt{Dialog}, \texttt{traj}}
\end{algorithm*}

\subsection{Dialogue Generation System}\label{dia}
\paragraph{Algorithmic Workflow of Multi-Agent Dialogue Generation}
Algorithm~\ref{alg} illustrates the pseudocode workflow of our multi-agent dialogue generation system, highlighting the interactions among different agents, the data flow between successive stages, and the overall generation and selection process.

\paragraph{Case Study}
We leverage a multi-agent framework to enable the dialogue generation system to produce three categories of dialogue: single-turn, multi-turn, and special cases. For both single-turn and multi-turn, we considered situations where users may accomplish either a single task or multiple tasks simultaneously. In addition, we incorporate special cases, such as when none of the available tools can address the user's request, or when the user's query prevents the model from filling in the tool parameters.

\begin{figure*}
    \centering
    \includegraphics[width=0.9\linewidth]{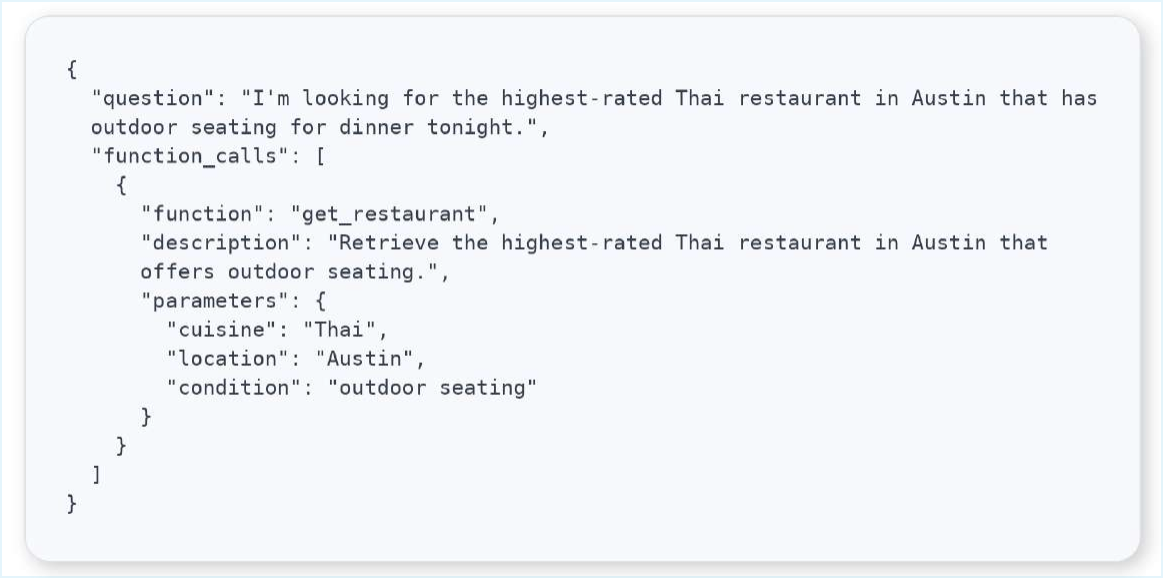}
    \caption{Example of a single-task scenario within single-turn dialogue.}
    \label{s-s}
\end{figure*}

\begin{figure*}
    \centering
    \includegraphics[width=0.9\linewidth]{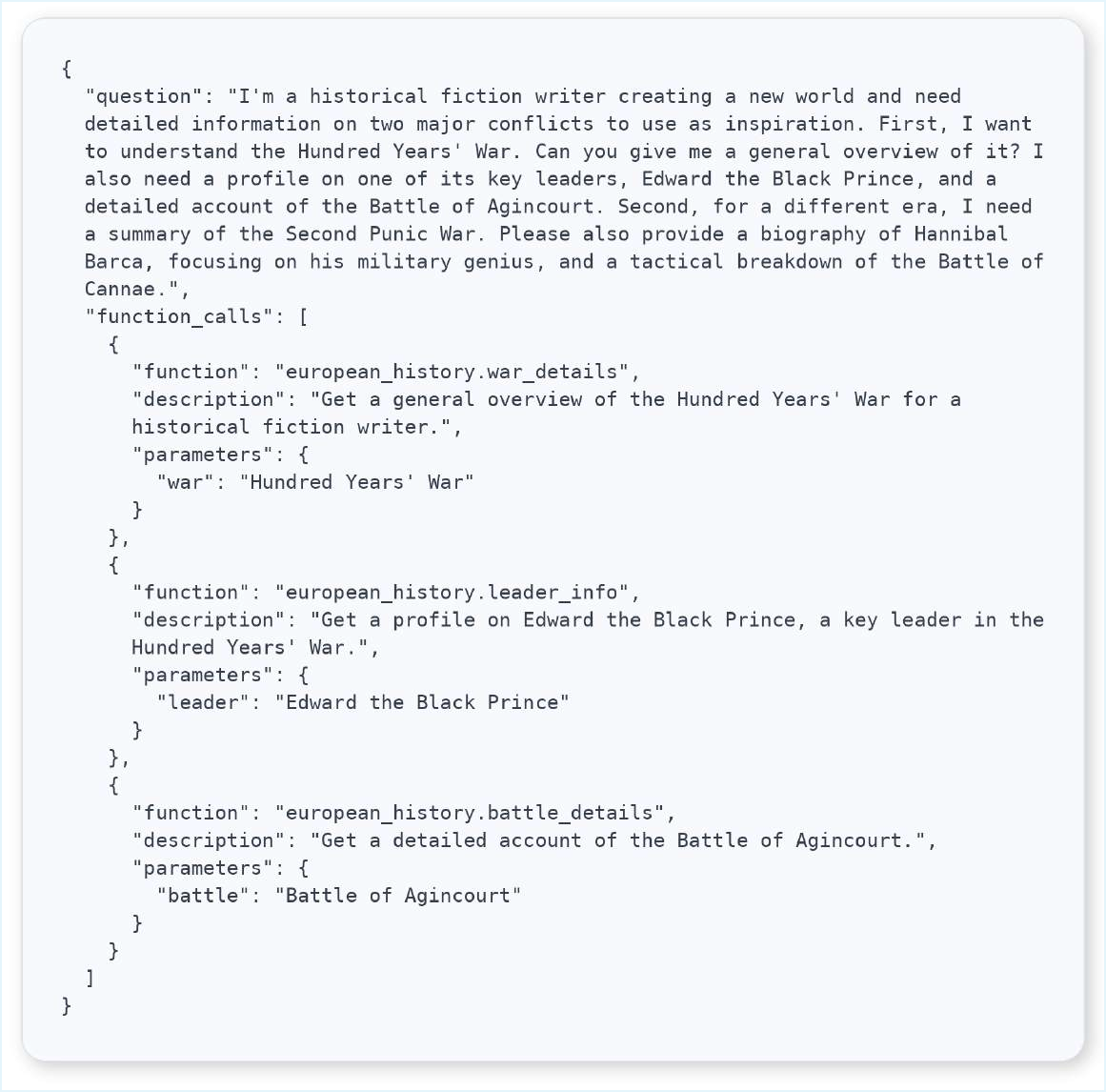}
    \caption{Example of a multi-task scenario within single-turn dialogue.}
    \label{s-m}
\end{figure*}

\begin{figure*}
    \centering
    \includegraphics[width=0.9\linewidth]{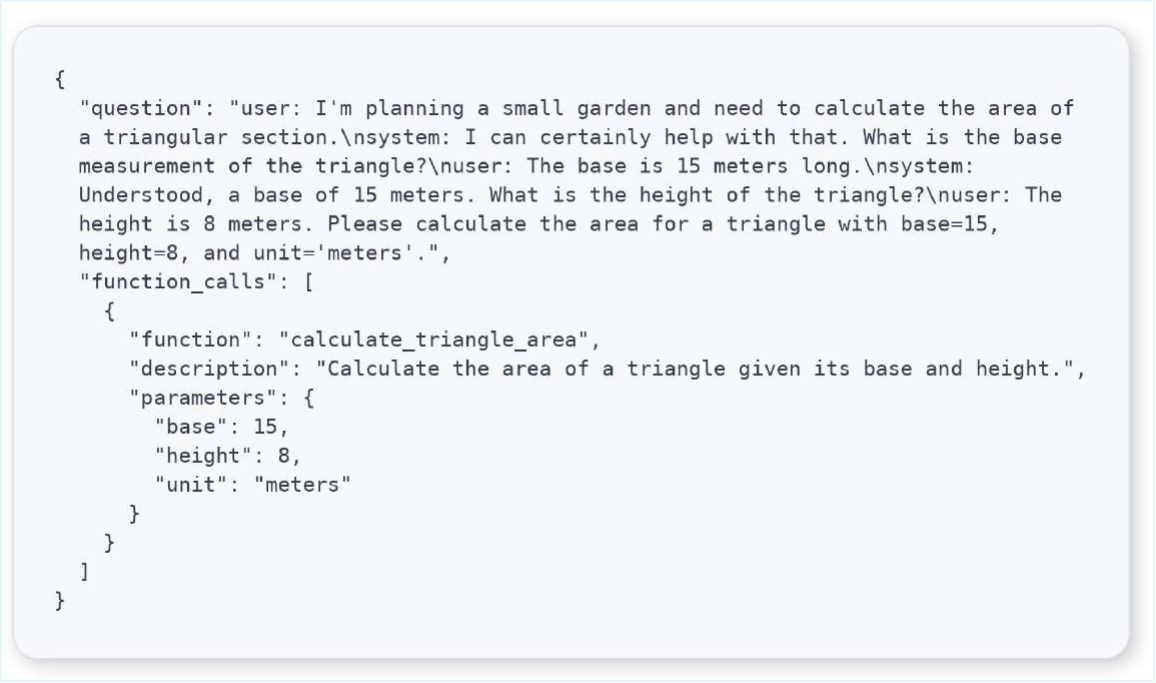}
    \caption{Example of a single-task scenario within multi-turn dialogue.}
    \label{m-s}
\end{figure*}

\begin{figure*}
    \centering
    \includegraphics[width=0.9\linewidth]{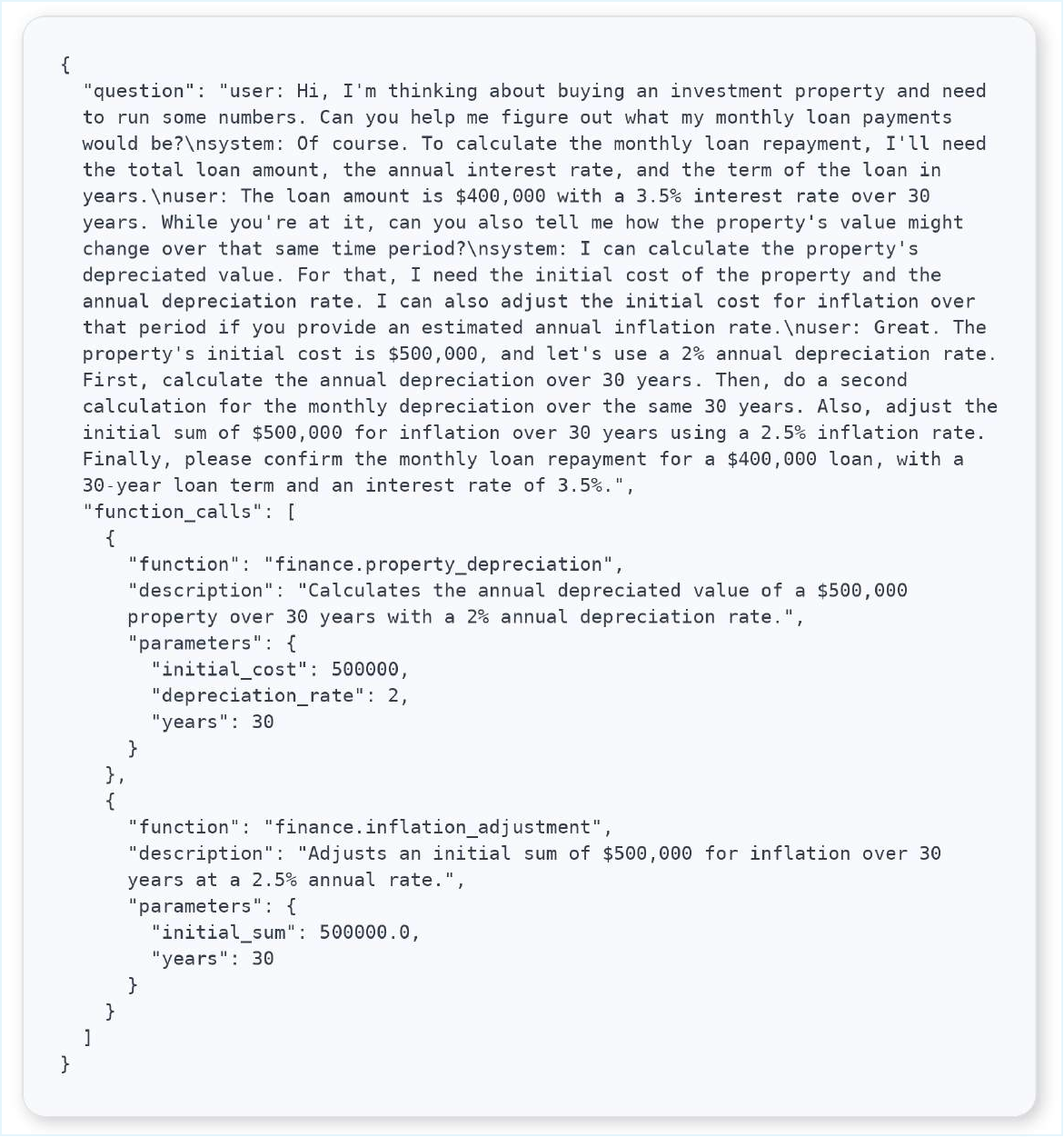}
    \caption{Example of a multi-task scenario within multi-turn dialogue.}
    \label{m-m}
\end{figure*}

\begin{figure*}
    \centering
    \includegraphics[width=0.9\linewidth]{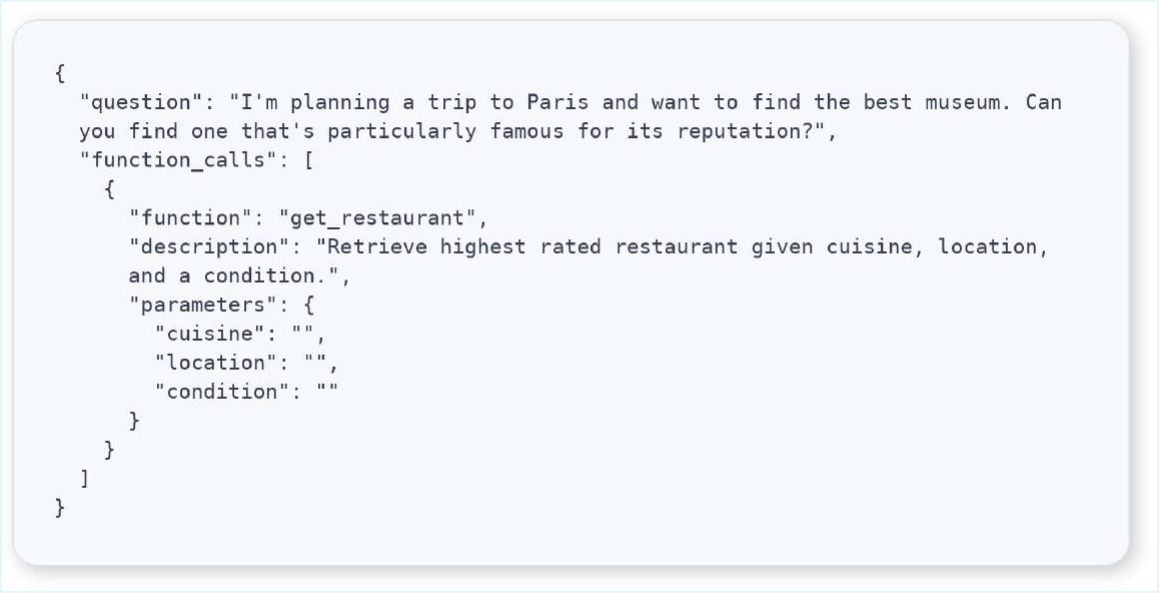}
    \caption{Example of a special-case dialogue.}
    \label{spe}
\end{figure*}

Figure~\ref{s-s} and Figure~\ref{s-m} illustrate single-turn scenarios in which the user intends to invoke one or multiple tools to complete either a single task or multiple tasks. Figure~\ref{m-s} and Figure~\ref{m-m} present analogous cases in multi-turn dialogues, where one or more tools are required to handle a single task or multiple tasks. Finally, Figure~\ref{spe} demonstrates the special-case dialogue that we constructed.

\subsection{Multi-Stage Evaluation}\label{app-rule}
\paragraph{Rule Checker}
Table~\ref{rule} outlines the check rules we use, which consist of four complementary aspects: tool definition completeness, call format and argument compliance, dialog structure soundness, and consistency between tool outputs and the assistant’s responses.

\paragraph{Model Checker}
The Model Checker verifies the correctness of function-calling dialogues using the following system-level prompt: \emph{``You are a Model Checker responsible for verifying the correctness of a function-calling dialogue. Given the dialogue, tool specifications, and tool-call outputs, determine whether the assistant selected the appropriate tool, used correct parameter formats, and provided a faithful answer. Identify any semantic errors and assign a confidence score between 0 and 1.''} 
This prompt defines the Model Checker’s role and behavior across evaluations.
We adopt GPT-4o~\cite{DBLP:journals/corr/abs-2303-08774} as the backend model due to its stable evaluative behavior, which improves robustness in automatic screening. This choice is an engineering consideration rather than an algorithmic requirement, and alternative models can be used without affecting the evaluation pipeline.

\begin{table*}[t]
\centering
\caption{Example rules for our Rule Checker in multi-stage evaluation.}
\label{rule}
\resizebox{\linewidth}{!}{%
\begin{tabular}{p{0.40\linewidth} p{0.56\linewidth}}
\toprule
\textbf{Aspect} & \textbf{Rules} \\
\midrule
\textbf{Tool Definition Completeness} &
Check that the tool schema parses and includes required fields (name, description, parameters). \\
& Verify that parameter specs declare type or format and required or optional flags. \\
& Ensure the tool name is unique within the tools pool. \\
\midrule
\textbf{Call Format \& Parameter Compliance} &
Confirm the called tools exist in the tools pool and use the correct name. \\
& Ensure all required parameters appear exactly once and no unknown keys exist. \\
& Validate that values satisfy declared type/regex/range constraints (e.g., ISO date, enum). \\
\midrule
\textbf{Dialogue Structure Soundness} &
Check that role order is valid. \\
& Ensure no dangling tool calls and every tool output is referenced by the assistant. \\
& Verify turn length within configured limits. \\
\midrule
\textbf{Tool-Assistant Consistency} &
Ensure assistant summaries faithfully reflect tool outputs without invented fields or values. \\
& Verify error codes and messages are propagated correctly. \\
& Check that tool names match the actual call. \\
\bottomrule
\end{tabular}%
} 
\end{table*}

\begin{table*}[t]
\centering
\caption{Comparison of dialogue scenario coverage across representative tool-augmented datasets. 
Here, \textit{Single-Single} denotes single-turn dialogues designed to accomplish a single task. 
\textit{Single-Multi} refers to single-turn dialogues that involve multiple tasks within the same interaction. 
\textit{Multi-Single} indicates multi-turn dialogues focused on a single task. 
\textit{Multi-Multi} represents multi-turn dialogues that cover multiple tasks. 
and \textit{Special-Case} corresponds to challenging scenarios such as the absence of applicable tools or cases where parameter filling is infeasible. Our dataset
uniquely covers all five categories.}
\label{compare}
\resizebox{\linewidth}{!}{%
\begin{tabular}{lccccc}
\toprule
Dataset & Single-Single & Single-Multi & Multi-Single & Multi-Multi & Special-Case \\
\midrule
Gorilla~\cite{DBLP:conf/nips/PatilZ0G24}      & \checkmark & $\times$ & $\times$ & $\times$ & $\times$ \\
ToolAlpaca~\cite{DBLP:journals/corr/abs-2306-05301}   & \checkmark & \checkmark & $\times$ & $\times$ & $\times$ \\
ToolLLM~\cite{DBLP:conf/iclr/QinLYZYLLCTQZHT24}      & \checkmark & \checkmark & $\times$ & $\times$ & $\times$ \\
xLAM~\cite{DBLP:conf/nips/LiuHZZLKTYLFNYS24}         & \checkmark & \checkmark & \checkmark & $\times$ & $\times$ \\
ToolACE~\cite{DBLP:conf/iclr/Liu0ZHYL0GLY0WN25}      & \checkmark & \checkmark & \checkmark & $\times$ & \checkmark \\
\textbf{\METHODNAME{}} & \checkmark & \checkmark & \checkmark & \checkmark & \checkmark \\
\bottomrule
\end{tabular}%
} 

\end{table*}

\section{Details of Training Data}\label{data}
Our constructed function-calling dataset consists of 2,000 single-turn dialogue scenarios, 2,000 multi-turn dialogue scenarios, and 500 special-case dialogues. To further demonstrate the strengths of our dataset in terms of quality, coverage, and diversity, we provide a set of statistical analyses.

\paragraph{Relationship Between Training Data and BFCL Evaluation}
Our training data uses the tool schemas provided by BFCL. During training, we do not use any BFCL test queries, test outputs, or instantiated evaluation samples. All dialogue are newly generated by our multi-agent framework, including user queries, dialogue contexts, and concrete parameter values.
Although the tool schemas are shared between training and evaluation, this setting follows the standard in-domain evaluation protocol of BFCL and prior function-calling benchmarks. The BFCL evaluation focuses on generalization to unseen queries and novel argument combinations rather than memorization of specific tool invocations. As a result, the observed performance gains reflect improved tool selection and parameter grounding instead of exposure to evaluation instances.

\paragraph{Quality}
In terms of quality, our constructed dataset features an average of 3.11 tools used per dialogue, 4.46 turns per multi-turn scenario, and 3.27 tasks per multi-task case. This design enables each dialogue to cover multiple tools, resulting in a substantially larger number of tool invocation instances than the number of dialogues, and the high number of tool calls, longer dialogue lengths, and richer task compositions expose the model to more complex tool-using situations, thereby better stimulating and enhancing its function-calling capability, while ensuring all data are newly generated and do not overlap with evaluation samples.

\paragraph{Coverage}
In terms of coverage, our dataset incorporates a broader range of scenarios compared to previously constructed dataset. This expanded scenario coverage is a key factor in strengthening the model's function-calling ability. An overview of the data statistics used in these representative tool-augmented LLMs is presented in Table~\ref{compare}.

\begin{figure}
    \centering
    \includegraphics[width=0.9\linewidth]{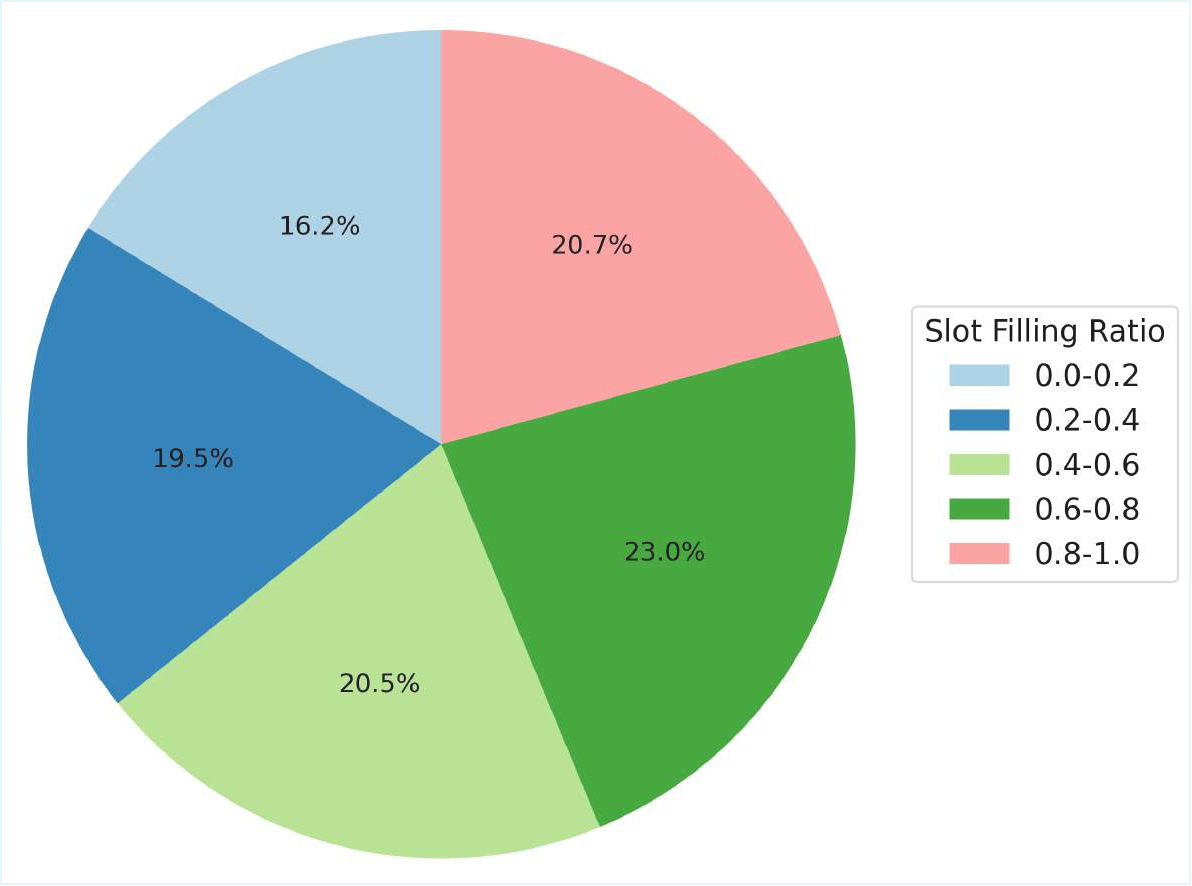}
    \caption{Distribution of slot filling ratios across 400 sampled tools.}
    \label{slot}
\end{figure}

\paragraph{Diversity}
In terms of diversity, we further analyze the slot filling ratios to better demonstrate the diversity of our tool-call training data. 
Specifically, we sample 400 tools with concrete parameter values from the constructed dialogues. 
For each tool, we compute the slot filling ratio by dividing the number of filled optional parameters by the total number of available optional parameters, and then group the results into five equal-width intervals. 
Figure~\ref{slot} reports the distribution across these ranges. 
The relatively balanced counts indicate that our dataset effectively captures heterogeneous slot utilization patterns, ranging from sparse filling to near-complete filling, highlighting its diversity.

\paragraph{Efficiency}
We examine the efficiency of the data construction process and compare generation cost with other synthetic datasets. Unlike synthetic function-calling datasets such as ToolLLM~\cite{DBLP:conf/iclr/QinLYZYLLCTQZHT24} and ToolACE~\cite{DBLP:conf/iclr/Liu0ZHYL0GLY0WN25} that rely on manually designed or synthesized APIs, our approach operates directly on the tools already defined in BFCL~\cite{BFCL}. This design avoids additional tool construction and manual schema engineering, thereby reducing the cost of dataset creation.
The generation cost of our approach arises from model generation within a multi-agent framework, while the tool space remains fixed and reusable across samples. Consequently, the overall construction cost scales linearly with the number of generated samples and does not introduce dataset-specific tool design or manual annotation overhead. Since existing datasets differ in tool complexity, agent design, and model backbones, a direct numerical comparison of generation cost is difficult. Nevertheless, under a unified benchmark setting, our approach achieves a balance between the quality of the training data and construction efficiency.

\section{Experimental Details}\label{app-exp}

\subsection{Benchmarks}
\paragraph{BFCL}
The Berkeley Function-Calling Benchmark (BFCL)~\cite{BFCL} is a comprehensive framework for evaluating the function-calling abilities of LLMs across multiple languages, domains, and complex scenarios. It includes 4,951 test cases, with 3,951 single-turn dialogs and 1,000 multi-turn dialogs that emphasize dynamic and realistic settings. 
The tools pool is broad, combining \textbf{Non-Live} tools curated to cover diverse situations with \textbf{Live} tools that are continuously uploaded by users. 
BFCL supports two evaluation methods, one based on Python and the other not. In this work, we adopt the Python-based approach, which is shown below:

\begin{itemize}
  \item \textbf{Simple Function:} This category represents the most basic yet also the most frequently encountered setting, where the input explicitly contains exactly one json function description and the model is expected to correctly invoke precisely that single function.
  
  \item \textbf{Multiple Function:} In this setting, the model receives 2 to 4 json function descriptions, but only one of them should be invoked. The task requires the model to identify and select the most suitable function call based on the user’s query and context.
  
  \item \textbf{Parallel Function:} Here, the model must invoke multiple function calls simultaneously in response to a single user query. The challenge lies in determining how many calls are required and executing them in parallel, whether the query is phrased as a short request or a longer description.
  
  \item \textbf{Parallel Multiple Function:} This combines the complexity of both multiple and parallel function settings. The model is given several function descriptions and must decide, for each one, whether it should be invoked, possibly multiple times or not at all.
\end{itemize}

For every category, BFCL provides both AST-based evaluation and a corresponding executable evaluation. In the executable setting, Python functions are manually implemented, inspired by publicly available REST API endpoints (such as retrieving weather data) as well as directly computable functions (such as linear regression). The purpose of this executable track is to assess whether the generated function calls can be reliably applied in real-world applications that depend on stable function execution.

\begin{table*}[h]
\centering
\caption{Configuration of hyper-parameters for model training.}
\label{hyper}
\resizebox{\linewidth}{!}{%
\begin{tabular}{ccccccc}
\hline
Learning Rate & Warmup Ratio & LR Scheduler & Batch Size & Epochs & LoRA Rank & LoRA Alpha \\ \hline
$10^{-4}$     & 0.1          & cosine       & 48         & 3      & 16        & 32         \\ \hline
\end{tabular}%
} 

\end{table*}

\paragraph{API-Bank}
The API-Bank~\cite{DBLP:conf/emnlp/LiZ000YLHL23} is composed of 314 dialogues involving a total of 753 API calls, specifically constructed to evaluate the abilities of LLMs in planning, retrieving, and invoking APIs in diverse scenarios. Within the dataset, 363 cases require only a single call, while 122 instances involve multiple calls, reflecting both simple and more complex usage patterns. Performance is systematically measured along three distinct dimensions, providing a comprehensive assessment of tool-use capability:  

\begin{itemize}
    \item \textbf{Call:} Evaluates whether the language model can correctly invoke a known API based on a given query.  
    \item \textbf{Retrieval+Call:} Evaluates the model's ability to first identify and accurately retrieve the correct API from context and then successfully perform the corresponding call when the API is not provided.  
    \item \textbf{Plan+Retrieval+Call:} Assesses the capacity to plan a sequence of actions, retrieve multiple APIs, and invoke them when the APIs are initially unknown.  
\end{itemize}  

The primary evaluation metric for API-Bank is \textit{accuracy}, formally defined as the ratio of correctly generated predictions to the total number of evaluation attempts.

\paragraph{ACEBench}
ACEBench \cite{DBLP:journals/corr/abs-2501-12851} is a bilingual Chinese and English benchmark for assessing LLMs tool use ability under realistic conditions. It contains about 2,000 annotated instances spanning a broad API set and organizes evaluation into three tracks:

\begin{itemize}
  \item 
  \textbf{Normal:} Single/multi-turn cases with similar-API and preference settings; calls are checked via AST matching against gold annotations.
  \item 
  \textbf{Special:} Inputs with missing, malformed, or otherwise irrelevant parameters are included to thoroughly test the model’s robustness when handling imperfect or noisy instructions.
  \item 
  \textbf{Agent:} Multi-turn, multi-step interactions in sandboxed scenarios, measuring process correctness and end-to-end task success.
\end{itemize}

Overall performance is computed from track-wise accuracies, providing fine-grained diagnostics of tool-use failures while avoiding reliance on live APIs or external LLM graders.

\subsection{Baselines}
Here we mainly introduce the two open-source models fine-tuned on function-calling data that we use as baselines, while details of the other open-source models and API-based models can be found in their publicly available technical reports.

\begin{itemize}
  \item 
  \textbf{ToolACE-8B}~\cite{DBLP:conf/iclr/Liu0ZHYL0GLY0WN25}: Obtained by fine-tuning LLaMA3.1-8B-Instruct with function-calling training data generated through data pipeline ToolACE.
  \item 
  \textbf{Qwen-ToolRL:} Obtained by fine-tuning Qwen3-8B with the publicly available function-calling training dataset ToolRL~\cite{DBLP:journals/corr/abs-2504-13958}, which is a hybrid corpus sampling 2K examples from ToolACE and 1K each from Hammer~\cite{DBLP:journals/corr/abs-2410-04587} and xLAM~\cite{DBLP:conf/naacl/ZhangLZLHKYTPCLFANCXNHW25}.
\end{itemize}

\subsection{Compute Cost}\label{CO}
We apply the parameter-efficient LoRA~\cite{DBLP:conf/iclr/HuSWALWWC22} strategy for fine-tuning and perform SFT using the LLaMA-Factory framework~\cite{DBLP:journals/corr/abs-2403-13372}. The primary computational overhead during generation arises from API calls, while post-training costs remain lightweight. Specifically, SFT is completed on a single A800 GPU in about 30 minutes, and the RL stage on four A800 GPUs in roughly 4 hours, making the overall compute requirements practical for reproduction.
The training data follows the Alpaca format, where the instruction includes the system prompt, tool pool, and user query, and the output contains the tool call. Hyper-parameters are shown in Table~\ref{hyper}.

\subsection{Statistical Significance and Variance Analysis}
All results reported in the main text are averaged over three runs with different random seeds. We also compute the standard deviation (reported as $\pm$SD in the tables) to assess model stability. Across all benchmarks, \METHODNAME{}-8B shows small standard deviations, typically below 0.5, indicating consistent performance across runs.
The observed performance gains of \METHODNAME{}-8B exceed twice the standard deviations on all benchmarks, demonstrating that the improvements are well beyond random variation.
To further verify robustness, we perform paired t-tests between \METHODNAME{}-8B and the strongest open-source baseline (ToolACE-8B), showing statistically significant gains (p < 0.05).

\begin{table*}[htbp]
\centering
\caption{Ablation study of different slot selection strategies in the function agent. To compare with our proposed dynamic slot selection approach, we also adopt two fixed strategies: filling all optional parameters and leaving them empty, and evaluate on BFCL. The best results are marked in bold, and the second best are \underline{underlined}.}
\renewcommand{\arraystretch}{1.25}
\resizebox{\textwidth}{!}{
\begin{tabular}{lcccccccccc}
\toprule
 & \multicolumn{5}{c}{\textbf{Non-Live}} & \multicolumn{5}{c}{\textbf{Live}} \\
\cmidrule(lr){2-6} \cmidrule(lr){7-11}
\textbf{Models} & \textit{Simple} & \textit{Multiple} & \textit{Parallel} & \textit{Parallel} & \textit{Overall} &\textit{Simple} & \textit{Multiple} & \textit{Parallel} & \textit{Parallel} & \textit{Overall}\\
 &  &  &  & \textit{Multiple} &  &  &  & & \textit{Multiple} \\
\midrule
\midrule
\textbf{Qwen3-8B} & 78.92 & 95.00 & 91.50 & 89.00 & 88.60 & 80.23 & 77.21 & \underline{81.25} & 75.00 & 77.79 \\
\textbf{ToolACE-8B} & 81.17 & 96.00 & 94.00 & \textbf{93.00} & 91.04 & 82.95 & 79.58 & 75.25 & \textbf{85.12} & 80.73\\
\textbf{\METHODNAME{}-8B(dynamic)} & \textbf{86.85} & \textbf{97.25} & \textbf{96.15} & \textbf{93.00} & \textbf{93.31} & \textbf{88.25} & \textbf{80.50} & \textbf{81.50} & \underline{84.88} & \textbf{83.78}\\
\textbf{\METHODNAME{}-8B(max)} & \underline{86.00} & 96.85 & 95.50 & \underline{92.25} & \underline{92.65} & 86.80 & \underline{80.00} & \underline{81.25} & 83.86 & \underline{82.98}\\
\textbf{\METHODNAME{}-8B(min)} & 85.75 & \underline{97.00} & \underline{95.75} & 92.00 & 92.63 & \underline{87.10} & 79.50 & 81.00 & 84.07 & 82.92\\
\bottomrule
\end{tabular}
}
\label{11}
\end{table*}

\section{Additional Experiments}
\subsection{Study on General Abilities}\label{gen}
\begin{figure}
    \centering
    \vspace{-2mm}    
    \includegraphics[width=0.9\linewidth]{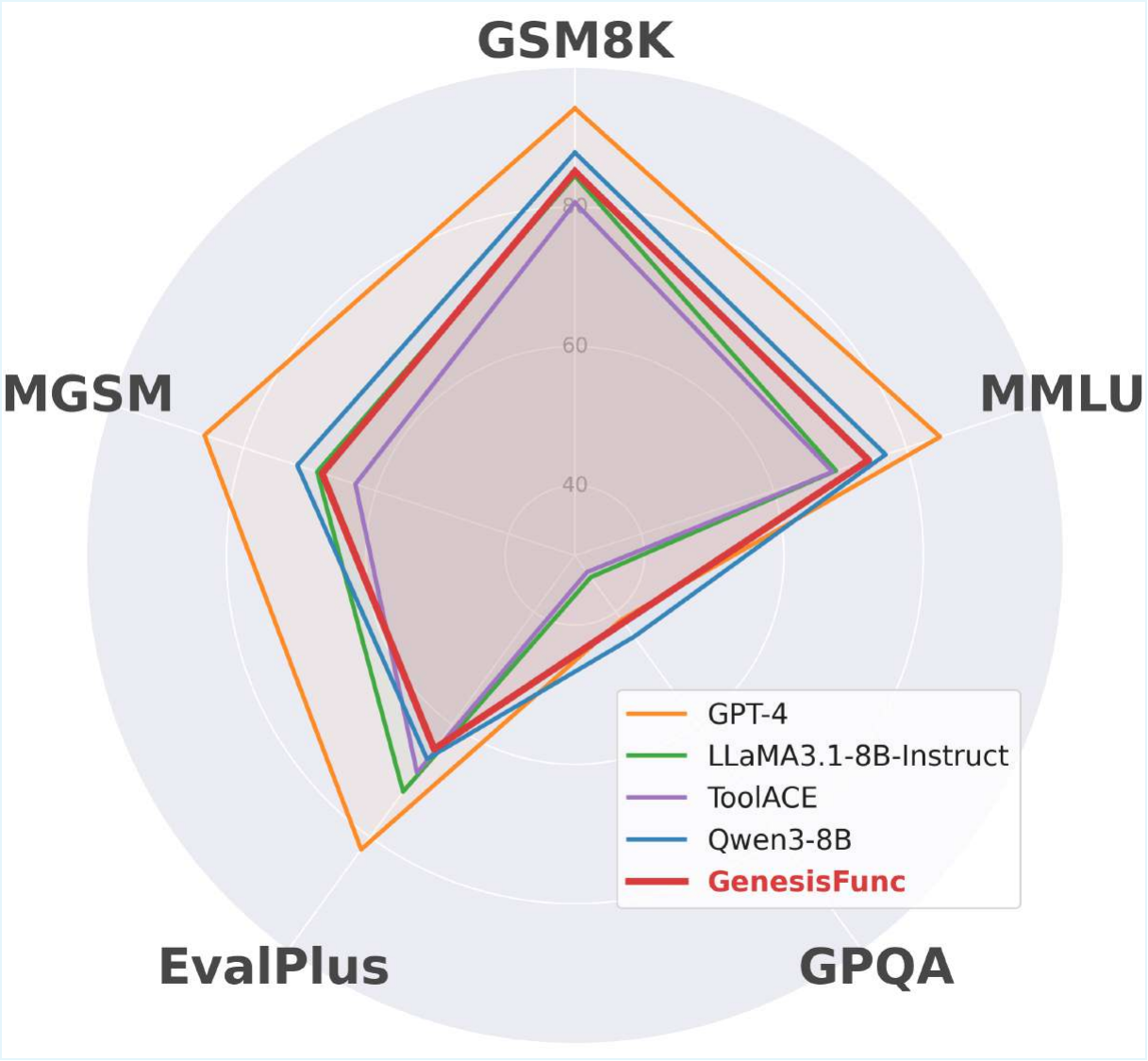}
    \caption{General abilities of models. Evaluation are conducted in five dimensions.}
    \label{general}
    \vspace{-7mm}
\end{figure}

To assess the broader impact of \METHODNAME{}-8B on general abilities, we evaluated it on five benchmarks: MMLU for knowledge~\cite{DBLP:conf/iclr/HendrycksBBZMSS21}, EvalPlus for code generation~\cite{DBLP:conf/nips/LiuXW023}, GSM8K for mathematics~\cite{DBLP:journals/corr/abs-2110-14168}, MGSM for multilingual reasoning~\cite{DBLP:conf/iclr/ShiSF0SVCTRZ0W23}, and GPQA for logical reasoning~\cite{DBLP:journals/corr/abs-2311-12022}. 
Baselines included LLaMA-3.1-8B-Instruct~\cite{DBLP:journals/corr/abs-2407-21783}, Qwen3-8B~\cite{DBLP:journals/corr/abs-2505-09388}, ToolACE-8B~\cite{DBLP:conf/iclr/Liu0ZHYL0GLY0WN25}, and GPT-4~\cite{DBLP:journals/corr/abs-2303-08774}. 
As shown in Figure~\ref{general}, \METHODNAME{}-8B performs on par with Qwen3-8B on most benchmarks, indicating that our fine-tuning substantially improves function-calling ability while preserving the model’s general abilities. It also surpasses similarly sized open-source models on many tasks, demonstrating competitive performance at the 8B scale. The remaining gap to GPT-4 in reasoning and comprehension is expected and likely stems from differences in model size and the breadth of training data rather than negative transfer from function-calling training. Overall, these results highlight the promise of targeted specialization for function-calling while maintaining broad competence, and they leave open the challenge of jointly improving multiple capabilities together with function-calling performance in a single model.

\subsection{Different Slot Selection Strategies in the Function Agent}\label{1}

Table~\ref{11} presents the results of different slot selection strategies in the function agent. Compared with two fixed strategies, filling all optional parameters (max) and leaving all optional parameters empty (min), our proposed dynamic slot selection approach achieves the best overall performance on both the Non-Live and Live settings. Specifically, \METHODNAME{}-8B(dynamic) reaches an overall accuracy of 93.31 on the Non-Live split and 84.83 on the Live split, surpassing both fixed strategies. 
These results demonstrate that dynamically selecting relevant slots enables the model to better capture diverse tool-use patterns while avoiding unnecessary noise, thereby improving the robustness and generalization of function-calling.

\subsection{Ablation on Multi-Stage Evaluation Module}\label{ABA}
As discussed earlier, we employ a multi-stage evaluation module to ensure the correctness of the dialogues generated in the previous stage. To evaluate its effectiveness, we fine-tune models on two datasets: one that has passed through this evaluation module and another that has not. The fine-tuned models are then evaluated on the BFCL benchmark, with results presented in Table~\ref{module}. The comparison clearly shows that models trained on data evaluated by the multi-stage module achieve higher overall accuracy than those trained on non-evaluated data, thereby demonstrating the effectiveness of our proposed evaluation module.

\begin{table*}[htbp]
\centering
\caption{Ablation study of multi-stage evaluation module. We remove the multi-stage evaluation module, and evaluate on BFCL. The best results in each category are
marked in bold. The second best results are
\underline{underlined}.}
\renewcommand{\arraystretch}{1.25}
\resizebox{\textwidth}{!}{
\begin{tabular}{lcccccccccc}
\toprule
 & \multicolumn{5}{c}{\textbf{Non-Live}} & \multicolumn{5}{c}{\textbf{Live}} \\
\cmidrule(lr){2-6} \cmidrule(lr){7-11}
\textbf{Models} & \textit{Simple} & \textit{Multiple} & \textit{Parallel} & \textit{Parallel} & \textit{Overall} &\textit{Simple} & \textit{Multiple} & \textit{Parallel} & \textit{Parallel} & \textit{Overall}\\
 &  &  &  & \textit{Multiple} &  &  &  & & \textit{Multiple} \\
\midrule
\midrule
\textbf{Qwen3-8B} & 78.92 & 95.00 & 91.50 & 89.00 & 88.60 & 80.23 & 77.21 & \underline{81.25} & 75.00 & 77.79 \\
\textbf{ToolACE-8B} & 81.17 & 96.00 & 94.00 & \textbf{93.00} & 91.04 & 82.95 & 79.58 & 75.25 & \textbf{85.12} & 80.73\\
\textbf{\METHODNAME{}-8B} & \textbf{86.85} & \textbf{97.25} & \textbf{96.15} & \textbf{93.00} & \textbf{93.31} & \textbf{88.25} & \textbf{80.50} & \textbf{81.50} & \underline{84.88} & \textbf{83.78}\\
\quad - \textbf{multi-stage evaluation module}   & \underline{85.17} & \underline{97.00} & \underline{96.00} & \underline{92.50} & \underline{92.68} & \underline{87.00} & \underline{80.00} & \underline{81.25} & 84.05 & \underline{83.08}\\
\bottomrule
\end{tabular}
}
\label{module}
\end{table*}

\begin{figure*}[t]
  \subfigure[Non-Live]{
  \includegraphics[width=0.47\textwidth]{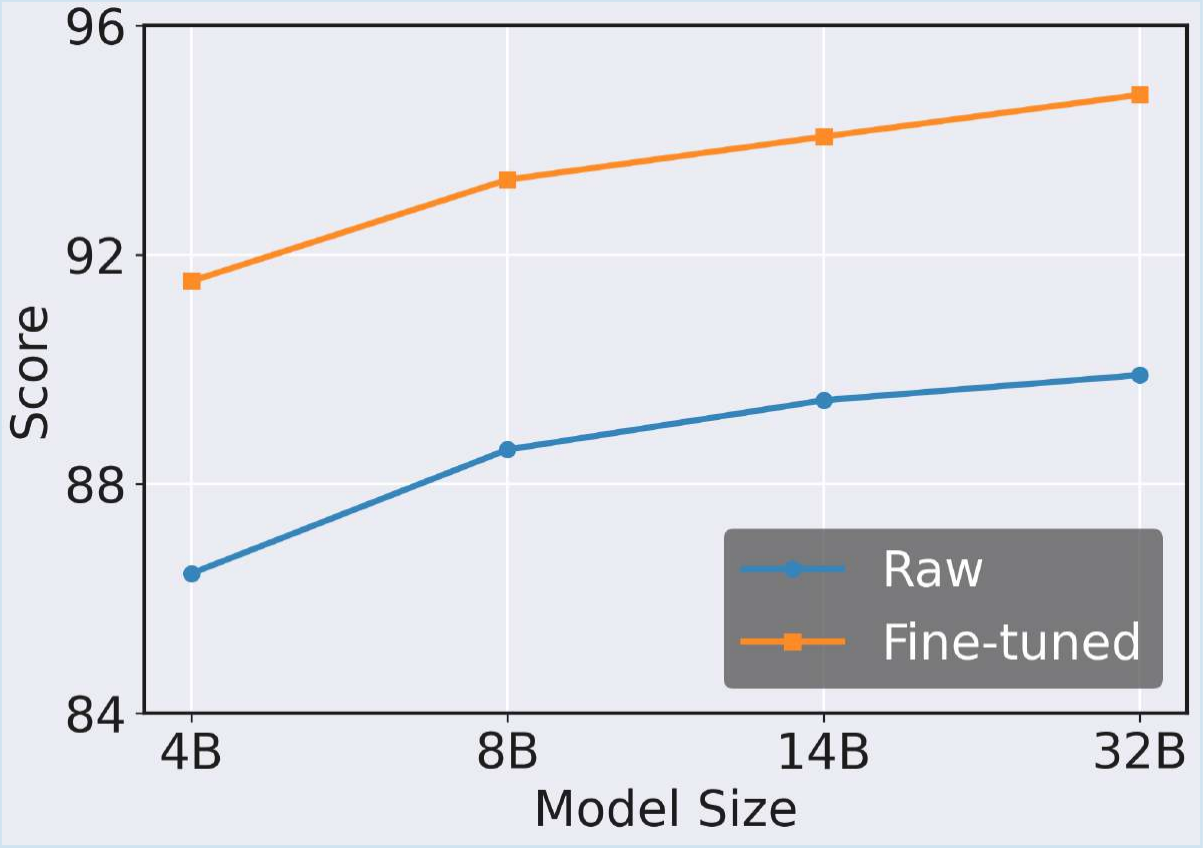}}
  \subfigure[Live]{
  \includegraphics[width=0.47\textwidth]{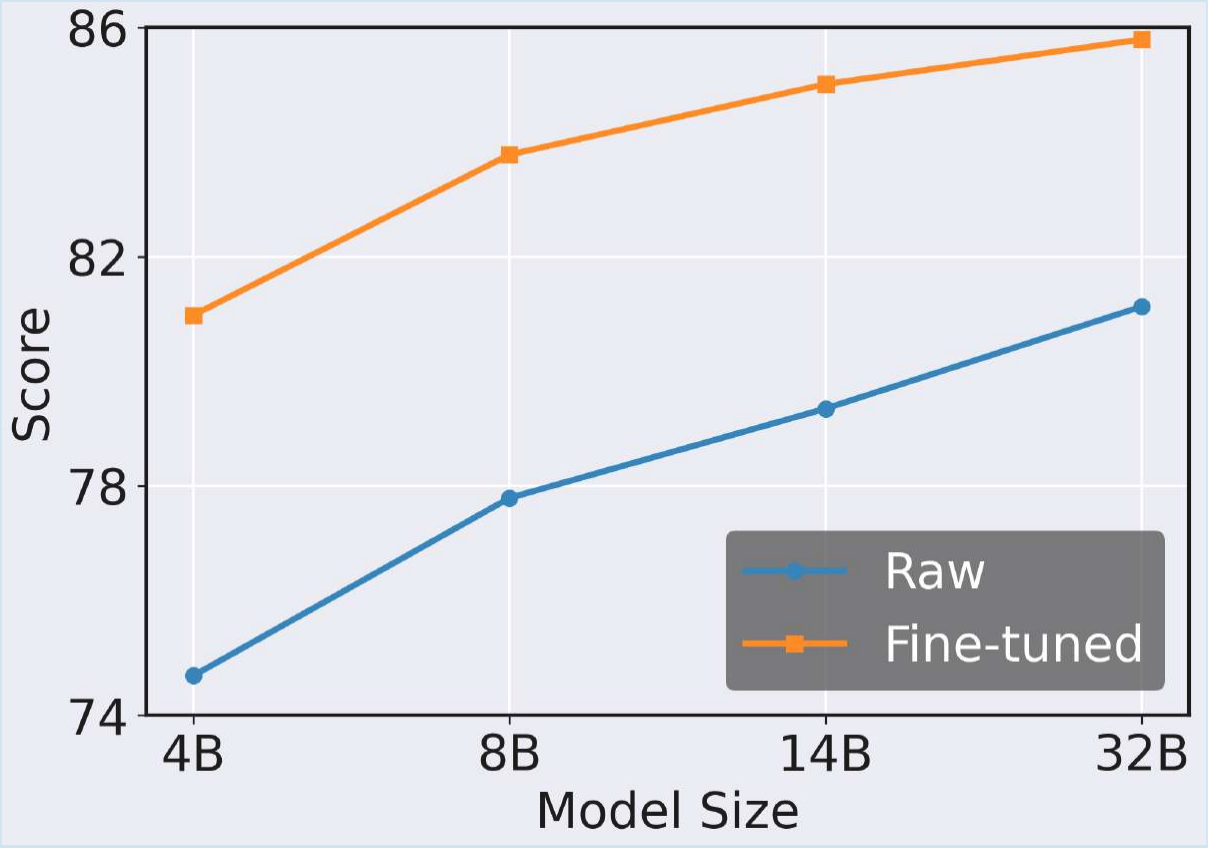}}
\vspace{-2mm}
\caption{Scaling analysis of function-calling performance. We evaluate raw and fine-tuned Qwen-3-xB models of different sizes on the BFCL benchmark in (a) Non-Live and (b) Live settings.}
\label{fig-scal}
\vspace{-5mm}
\end{figure*}

\subsection{Scaling to Different Model Sizes}\label{scale}
Scaling laws suggest a close relationship between model capacity and empirical performance. To examine how function-calling ability scales with model size, we evaluate the Qwen-3-xB model family~\cite{DBLP:journals/corr/abs-2505-09388}, covering a range of parameter scales (4B, 8B, 14B, and 32B). Both raw models and models fine-tuned on our dataset are assessed on the BFCL~\cite{BFCL}, and the corresponding results are presented in Figure~\ref{fig-scal}.
Overall, the results show a steady improvement in function-calling performance as model size increases across both non-live and live evaluation settings, with larger models exhibiting stronger robustness and more consistent execution accuracy. In comparison to raw models, fine-tuned models consistently achieve higher performance at all scales, indicating that supervision from the \METHODNAME{} effectively enhances function-calling behavior. Importantly, the fine-tuned models maintain a smooth and stable scaling pattern, suggesting that \METHODNAME{} complements model capacity rather than altering the underlying scaling dynamics. Together, these observations demonstrate the effectiveness of \METHODNAME{} in supporting scalable performance gains for LLMs.

\begin{figure*}[t]
  \subfigure[Non-Live]{
  \includegraphics[width=0.47\textwidth]{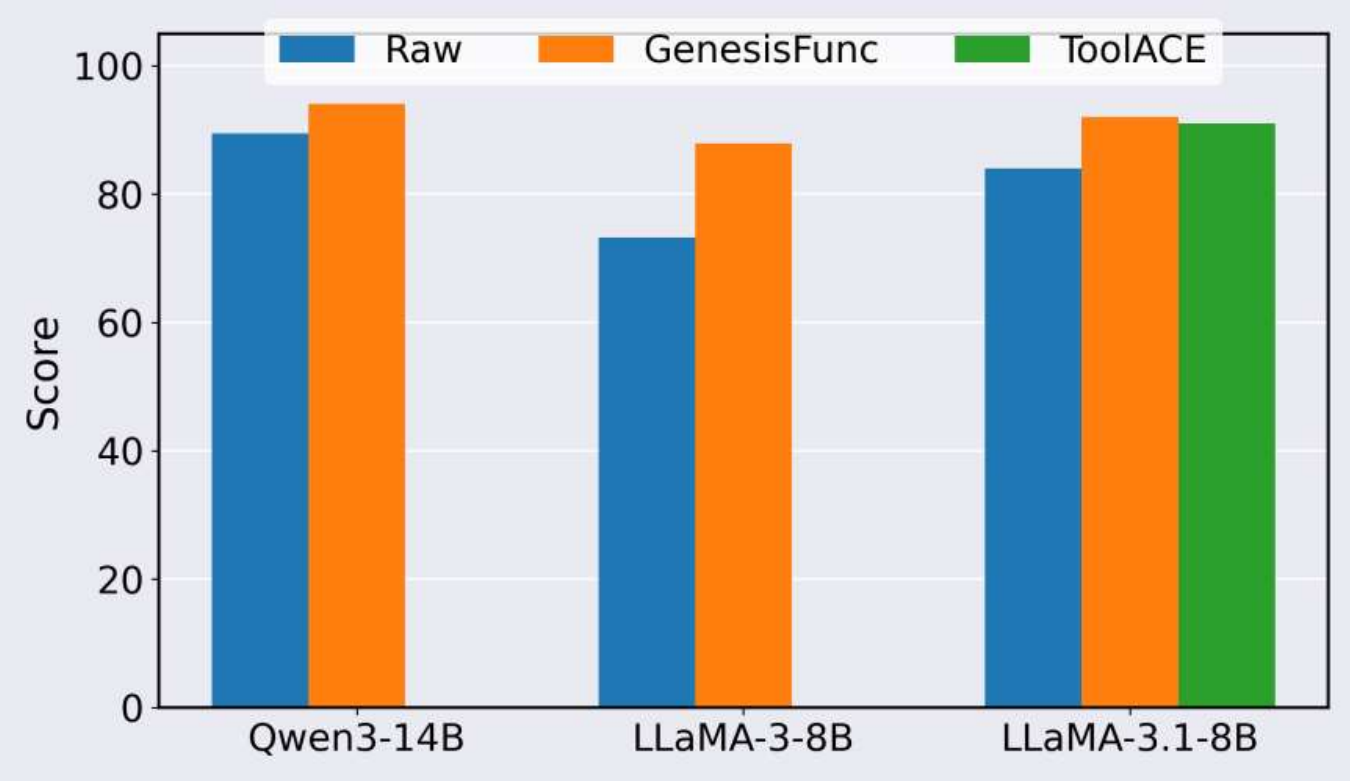}}
  \subfigure[Live]{
  \includegraphics[width=0.47\textwidth]{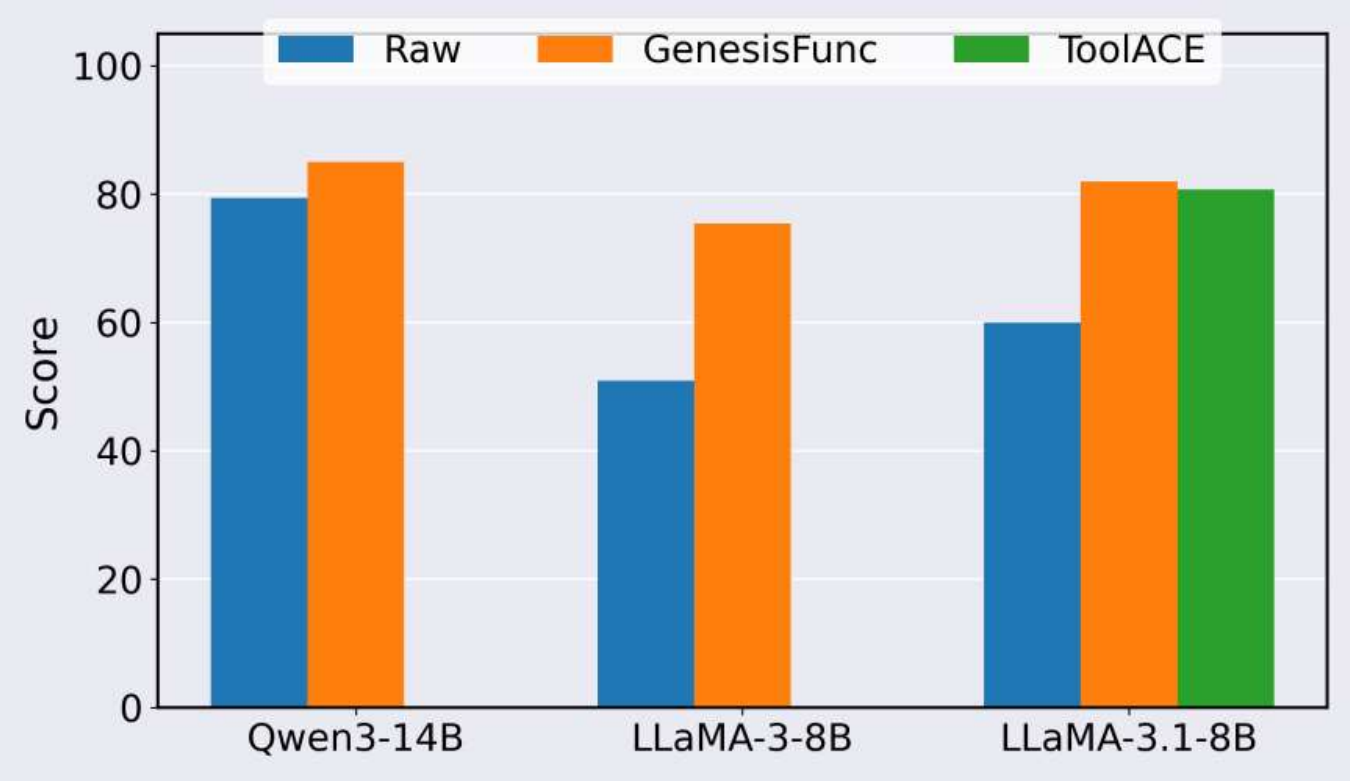}}
\vspace{-2mm}
\caption{Generalization across model backbones. We evaluate raw and fine-tuned LLMs with different backbone architectures on the BFCL benchmark in (a) Non-Live and (b) Live settings.}
\label{fig-backbone}
\vspace{-5mm}
\end{figure*}

\subsection{Generalization across Model Backbones}\label{backbone}
To analyze the impact of backbone choice on function-calling performance, we evaluate a set of representative LLMs, including Qwen3-14B, LLaMA-3-8B-Instruct, and LLaMA-3.1-8B-Instruct. All models are fine-tuned using our dataset and evaluated on the BFCL benchmark. In addition, to mitigate potential confounding effects introduced by different backbones and to more rigorously isolate the contribution of the data-generation pipeline itself, we conduct a controlled comparison on the same backbone. Specifically, since ToolACE is trained on LLaMA-3.1-8B-Instruct, we directly compare the two data-generation pipelines on this backbone. The comparative results are summarized in Figure~\ref{fig-backbone}.
Figure~\ref{fig-backbone} shows that fine-tuning with \METHODNAME{} consistently improves performance across all examined backbones under both non-live and live evaluation settings, demonstrating strong cross-backbone robustness. Despite differences in model architectures and pre-training objectives, fine-tuning yields stable and significant gains overall. Notably, models with lower initial function-calling accuracy tend to exhibit larger relative improvements, thereby narrowing the performance gap across backbones. More importantly, the controlled comparison on LLaMA-3.1-8B-Instruct indicates that, under the same backbone, \METHODNAME{} outperforms ToolACE, providing stronger evidence that the observed gains stem from the data-generation pipeline rather than backbone-specific characteristics.

\section{Details of Reinforcement Learning}\label{RL}

\subsection{GRPO Algorithm}
To fine-tune the model with structured rewards, we adopt \textit{Grouped Relative Policy Optimization} (GRPO), a variant of PPO that normalizes advantages within groups of responses derived from the same input query. This design mitigates variance across samples under identical contexts, thereby stabilizing and accelerating training. Specifically, for each query $Q$, the rollout responses form a group $G_Q = \{(s_1,r_1), (s_2,r_2), \ldots, (s_n,r_n)\}$, where $s_i$ denotes a candidate response and $r_i$ its reward. Each reward is obtained as the sum of correctness and formatting scores relative to the reference annotation. Within each group, the mean $\mu_Q$ and standard deviation $\sigma_Q$ of rewards are computed as:
\begin{equation}
   \mu_Q = \frac{1}{n}\sum_{i=1}^n r_i, \quad 
\sigma_Q = \sqrt{\frac{1}{n}\sum_{i=1}^n (r_i - \mu_Q)^2}, 
\end{equation}
and the normalized advantage of response $s_i$ is defined as:
\begin{equation}
A(s_i|Q) = \frac{r_i - \mu_Q}{\sigma_Q + \eta},
\end{equation}
where $\eta$ is a small constant added for numerical stability.

The policy $\pi_\theta$ is then updated using the clipped PPO objective with group-normalized advantages:
\begin{equation}
r_\theta(s_i|Q)=\tfrac{\pi_\theta(s_i|Q)}{\pi_{\text{old}}(s_i|Q)}.
\end{equation}

\begin{equation}
\begin{aligned}
J_{\text{GRPO}}(\theta)
= \mathbb{E}\big[
\min(
& r_\theta(s_i|Q)\,A(s_i|Q), \\
& \operatorname{clip}\big(
r_\theta(s_i|Q),\,
1-\epsilon,\,
1+\epsilon
\big) \\
& \qquad\qquad \times A(s_i|Q)
)
\big].
\end{aligned}
\end{equation}
Unlike standard PPO, GRPO omits the KL penalty against a reference model, thereby allowing greater flexibility in adapting to diverse custom reward functions while still retaining training stability. This modification improves overall sample efficiency, encourages stronger alignment with structured reward signals, and leads to faster convergence as well as more robust policy performance.

\subsection{Reward Design}
Reward functions play a central role in reinforcement learning, guiding models to generate outputs that are both valid and useful. Following prior studies on rule-based reward signals~\cite{DBLP:journals/corr/abs-2502-14768, DBLP:journals/corr/abs-2503-09516, DBLP:journals/corr/abs-2503-23383, DBLP:journals/corr/abs-2504-13958}, we adopt a two-dimensional design that combines structural compliance with functional correctness, tailored to the demands of tool-augmented dialogue.

\paragraph{Structural Compliance}
To ensure that generated outputs conform to the expected schema, we introduce a \textit{structural compliance reward}. This reward evaluates whether each predicted tool call follows the prescribed format, including the presence of all mandatory fields and the correct logical ordering of elements. For single-tool cases, the reward reduces to a simple binary check defined as
\begin{equation}
r_{\text{structural}}^{(i)} =
\begin{cases}
1, &
\begin{aligned}
&\text{if the $i$-th output follows} \\
&\text{the required schema},
\end{aligned} \\
0, & \text{otherwise}.
\end{cases}
\end{equation}
For multi-tool cases, the score is computed for each tool call individually according to the above rule and then averaged across all calls:
\begin{equation}
R_{\text{structural}} = \frac{1}{N} \sum_{i=1}^{N} r_{\text{structural}}^{(i)},
\end{equation}
where $N$ denotes the number of tools in the output.

\paragraph{Functional Correctness}
In addition to structural validity, we further assess whether the predicted tool calls can successfully achieve the intended functionality. For this purpose, we introduce a \textit{functional correctness reward}. This reward explicitly captures the degree of semantic and functional agreement between the predicted calls and the reference calls. For a single-tool case, the reward is defined as follows:
\begin{equation}
r_{\text{correct}}^{(i)} =
\begin{cases}
3, &
\begin{aligned}
&\text{if the tool name and all parameters} \\
&\text{match exactly},
\end{aligned} \\
2, &
\begin{aligned}
&\text{if the tool name is correct and} \\
&\text{at least one parameter matches},
\end{aligned} \\
1, & \text{if only the tool name is correct}, \\
0, & \text{otherwise}.
\end{cases}
\end{equation}
For multi-tool cases, the correctness score is first computed for each tool call using the above rule, and then the results are averaged across all calls:
\begin{equation}
R_{\text{correct}} = \frac{1}{N} \sum_{i=1}^{N} r_{\text{correct}}^{(i)},
\end{equation}
where $N$ denotes the number of tools in the output.

Finally, we combine the two components to form the overall reward function. The structural compliance reward ensures that outputs remain syntactically valid, while the functional correctness reward encourages accurate execution of tool calls. By integrating both signals, the model receives feedback that jointly emphasizes format validity and semantic accuracy. The final reward is simply defined as follows:
\begin{equation}
R_{\text{final}} = R_{\text{structural}} + R_{\text{correct}}.
\end{equation}
This unified formulation provides a balanced training signal, preventing the model from generating structurally invalid outputs and at the same time guiding it toward functionally reliable tool usage.

\end{document}